# Transformer Based Model for Predicting Rapid Impact Compaction Outcomes: A Case Study of Utapao International Airport

Sompote Youwai[1*]
Sirasak Detcheewa[2]

[1]Associate Professor
[2]Master Student
Department of Civil Engineering, King Mongkut's University of Technology Thonburi, Bangkok, Thailand
[*]Corresponding author- Email: sompote.you@kmutt.ac.th

**Abstract**
This paper introduces a novel deep learning approach to predict the engineering properties of the ground improved by Rapid Impact Compaction (RIC), which is a ground improvement technique that uses a drop hammer to compact the soil and fill layers. The proposed approach uses transformer-based neural networks to capture the complex nonlinear relationships between the input features, such as the hammer energy, drop height, and number of blows, and the output variables, such as the cone resistance. The approach is applied to a real-world dataset from a trial test section for the new apron construction of the Utapao International Airport in Thailand. The results show that the proposed approach outperforms the existing methods in terms of prediction accuracy and efficiency and provides interpretable attention maps that reveal the importance of different features for RIC prediction. The paper also discusses the limitations and future directions of applying deep learning methods to RIC prediction.
**Keywords:** Rapid impact compaction, Deep learning, Transformer

## 1. Introduction

Rapid impact compaction (RIC) is a method that uses medium-energy mechanical impacts to increase the soil engineering properties by compacting it. It is often used in large infrastructure projects such as airports and highways, where the soil needs to support the weight of the structure and pavement (Cheng et al. 2021; Mohammed et al. 2013; Simpson et al. 2008; Spyropoulos et al. 2020; Tarawneh and Matraji 2014; Vukadin 2013). The effectiveness of RIC depends on various factors, such as the fine content of the soil, the compaction sequence, the energy applied, the stiffness of existing ground, the ground water characteristics and the soil drainage. These factors vary in different site conditions and need to be considered in the design of RIC to optimize the compaction method (Ghanbari and Hamidi 2014; Serridge and Synac 2006; Tarawneh and Matraji 2014). Therefore, it is recommended to conduct a trial before the actual construction.

Predicting the engineering properties of the ground improved by Rapid Impact Compaction (RIC) is a challenging task for geotechnical engineers. Several factors influence the effectiveness of RIC, such as the soil profile, the engineering properties of fill, and the number of blows. Typically, RIC projects start with field trial tests to obtain the initial parameters for designing the RIC construction sequence in a real field construction. However, the results from the trial tests of RIC may not reflect the actual conditions of the construction site, where the soil profile and other conditions may vary. The change in soil profile may affect the compaction performance and the outcome of RIC. Therefore, a model of RIC to predict its behavior is strongly required for design and supervision stages. However, currently there is no such model available. The input data for predicting the outcome of RIC is a sequential data composed of discrete properties of soil layers obtained from the trial tests. A possible approach to model the behavior of RIC is to apply deep learning methods for sequential data analysis.

Long short-term memory (LSTM) (Hochreiter and Schmidhuber 1997; Van Houdt et al. 2020) and convolutional neural network (CNN) (Fukushima 1980) are two prevalent deep learning models for various tasks involving sequential and spatial data. LSTM is a recurrent neural network that can model the long-term dependencies and temporal dynamics of sequential data, such as natural language, speech, and time series. CNN is a feedforward neural network that can learn the local and global features of spatial data, such as images, videos, and graphs. Both models have achieved remarkable performance in domains such as natural language processing, computer vision, speech recognition, and bioinformatics. However, some tasks require simultaneous processing of both sequential and spatial information, such as video analysis, sentiment analysis, and human activity recognition. In these cases, a hybrid model that integrates LSTM and CNN can be more effective than using either model separately. A hybrid LSTM-CNN model can leverage the advantages of both models: the CNN layer can extract the spatial features from the input data, while the LSTM layer can capture the temporal dependencies between the extracted features. The hybrid model can also reduce the computational complexity and the number of parameters compared to using a single model with a large architecture. Several studies have proposed and applied hybrid LSTM-CNN models for different tasks (Alhussein et al. 2020; Khatun et al. 2022; Sagnika et al. 2021). The hybrid LSTM-CNN model is a promising deep learning framework that can handle complex tasks involving both sequential and spatial data.

The Transformer architecture is a prominent and effective deep learning approach for generative modelling in various domains, such as text-to-text ( Devlin et al. 2019; OpenAI 2023; Touvron et al. 2023) text-to-image (Ding et al. 2021) and image to text generation (Wang et al. 2021a; Wei et al. 2021). The key innovation of this architecture is the use of attention mechanisms (Vaswani et al. 2017), which enable a neural network to focus on relevant parts of the input or output depending on the task. This technique mimics the human attention process, which filters out irrelevant information and enhances the learning of pertinent information. Attention mechanisms are the basis for large-scale pre-trained language models, such GPT-4 (OpenAI 2023), BERT, and LLaMa (Touvron et al. 2023)which have achieved remarkable results in natural language processing tasks. The Transformer architecture consists of an encoder and a decoder, which are connected by latent variables. The encoder transforms the input words into tensors with positional embeddings, and then applies an attention-based correlation process to generate query responses. The decoder uses these responses to produce the output words. We hypothesize that the Transformer model architecture could be applied to a regression problem in the domain of discrete soil property information, such as borehole data or field tests like Cone Penetration Testing (CPT). By embedding engineering-specific properties into the tensors, the Transformer model can predict desired geotechnical outcomes. However, there is limited research on using Transformer-based generative models as predictive models in geotechnical engineering. The trained deep learning model can be used as a generative model similar to large language models. The generative predicted value can be obtained by inputting the initial soil profile and feature. This might be useful for geotechnical projects in design and supervision to predict the unforeseen value prior and during construction.

This paper proposes a novel neural network approach for generative modeling of rapid impact compaction (RIC) outcomes using data from the trial test section for the new apron construction of the Utapao International Airport in Thailand. RIC performance prediction is challenging due to the complex interactions among soil properties, compaction effort, and fill thickness. The model types that we evaluate in this paper were feed forward and sequence to sequence deep learning (Seq2Seq) architecture. For feed forward type model, this research evaluates different types of neural network architectures, such as fully connected (Dense), convolutional neural network (CNN), and long short-term memory (LSTM), for their prediction accuracy. Then, a hybrid model between LSTM and CNN is developed and compared with the other models. In the second part, a sequence to sequence model (Sutskever et al. 2014, Vaswani et al. 2017) based on the transformer architecture with attention mechanism is adopted to predict the RIC outcomes in a sequential manner. The modification of the transformer sequence to sequence model with including recurrent neural network into model was proposed in this study as triad hybrid model. The proposed model incorporates the compaction effort, fine content in compacted soil, thickness of fill material, and initial soil profile as input variables and predicts the RIC test results as output variables. The last part of this paper present the application of the trained generative model and parametric study with varying compaction effort. The paper also discusses how to train the generative model and its limitations and future directions. The main contributions of this study are:
- We develop a comprehensive generative model based on transformer architecture of RIC compaction that incorporates various factors affecting the RIC performance, such as compaction effort, fine content, fill thickness, and initial soil profile.
- We evaluate the performance of our proposed model against conventional deep learning methods using a real-world dataset from the trial test section for the new apron construction of the Utapao International Airport in Thailand.

## 2. Test section
### 2.1 Overview of project

Thailand. Previously, this airport belonged to the Royal Thai Navy and served as a base for the US Air Force during the Vietnam War. The Thai government planned to expand the capacity of the airport to accommodate the industrial estate nearby and the tourist destination known as Thailand Eastern Economic Corridor (EEC). The site location was near the sea with a beach nearby. The site terrain of the airport had a high elevation that gradually decreased towards the sea. The location of the new passenger terminal was 3-5 m lower than the expected design level. Therefore, the existing ground had to be filled with fill material. The compaction of fill material was a critical issue for the construction due to the large area of compaction, up to millions of m$^2$. The method of construction and material selection was determined by conducting a trial test on the construction site. The rapid impact compaction was chosen as the most suitable method for constructing the fill apron.

### 2.2 Rapid impact compaction machine

Soil compaction with a Rapid Impact Compactor (RIC) machine was performed using a 90-kN drop hammer and a 1.5-m diameter circular plate. The plate contacted the ground and applied stress to the low-stiffness soil, increasing its density. The compaction process, illustrated in Figure 1, resembled the laboratory compaction test with a hammer. The machine delivered 50 blows per minute to the soil, taking 1-2 minutes per location. The plate spacing and compaction effort were varied to achieve the target field density and optimize the construction cost and time. The literature reported that the hammer weight ranged from 50 to 120 kN and the drop height was about 1.2-1.5 m(Cheng et al. 2021; Serridge and Synac 2006; Tarawneh and Matraji 2014; Vukadin 2013). A higher drop height increased the compaction energy but also the compaction time. The soil and site conditions affected the selection of these parameters. Excessive compaction effort could create deep craters or holes that could destabilize the machine. The optimal energy for compaction was determined as described in the next section.

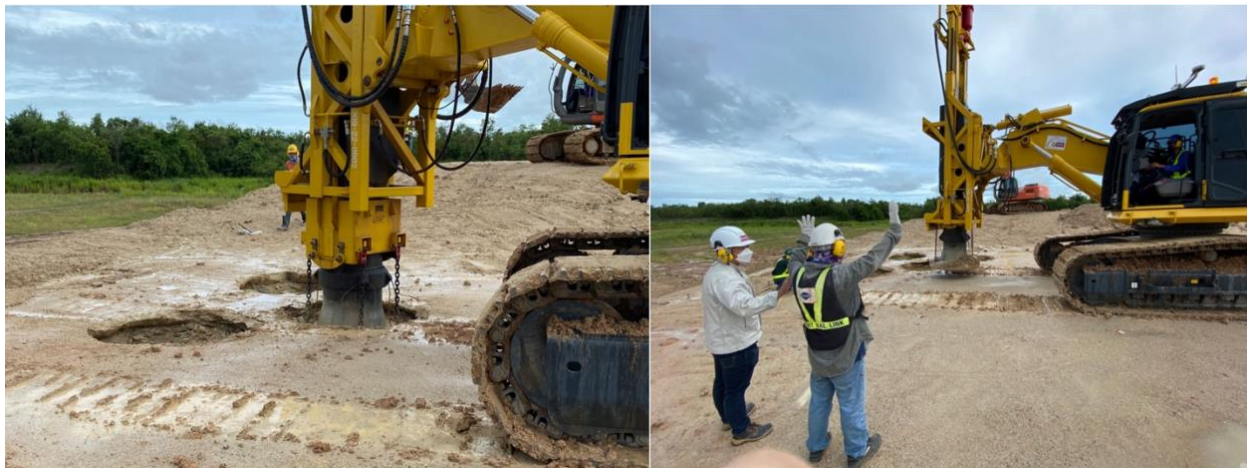

**Figure.1** The RIC machine

### 2.3 Geotechnical characteristics of the construction site

The site-specific ground conditions posed significant challenges for the application of Rapid Impact Compaction (RIC) in this construction project (Fig. 2). The site was located near the sea and the soil was mostly classified as silty sand (SM) according to the Unified Soil Classification System (USCS). The soil consisted of medium dense sand with high fine content (30-40 %), which was unfavorable for RIC. Moreover, the shallow groundwater table (2 m from the ground surface) and the presence of a weak silt layer at some locations increased the difficulty of RIC. The silt layer at 2 m depth could experience punching failure due to the RIC process. Therefore, a compacted layer platform had to be established above the existing ground to prevent punching shear failure and water from emerging from the compaction crater. The upper layer (0-1 m) had a higher stiffness than the lower layer, as indicated by the cone penetration test (CPT) results. The cone resistance value was high near the ground surface and decreased with depth. The minimum cone resistance value was about 2 MPa at 2.8 m depth and increased with further depth. The RIC technique could enhance the density and stiffness of sand by applying dynamic loads to the ground surface.

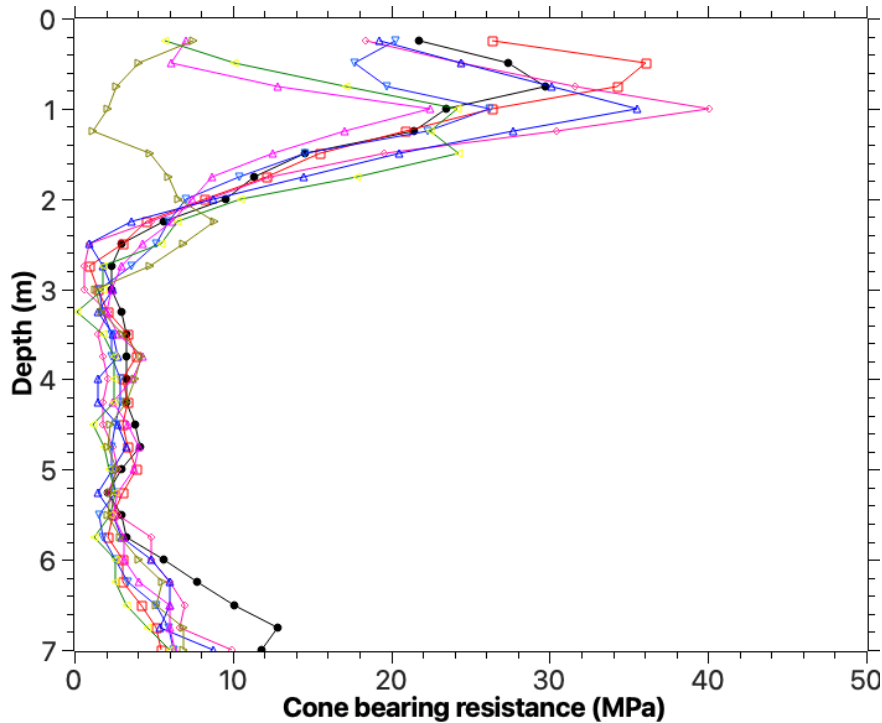
**Figure 2** Cone resistance at the original area.

**2.4 Test Embankment**

To improve the ground compaction and develop a construction method for fill embankment in Apron, a full-scale test embankment was conducted (Table 1). The existing ground level was lower than the expected apron level, so a fill embankment with a maximum height of 5 m from the surface was planned. The Rapid Impact Compaction (RIC) method was used to compact the fill embankment, as it can achieve an effective compaction depth of 3 to 5 m depending on the fine content of the soil (Cheng et al. 2021; Mohammed et al. 2010; Spyropoulos et al. 2020). To reduce the cost of fill material, two scenarios were tested: one using soil cut from the construction site with high fine content (SC), and the other using soil imported from outside with fine content lower than 20% (SM). The expected compaction depth was varied from 3 to 5 m. The design team hypothesized that the compaction would be effective for 3 m based on previous case studies, but they also wanted to investigate the possibility of increasing the fill thickness to 5 m. Another scenario was to compact the existing ground, which had a high groundwater level and low stiffness. A fill thickness of 0.5 m was used as a platform to prevent punching shear to the low strength existing soil layer. The feasibility of using SC as fill material was evaluated as the objective.

The experimental design for the full-scale test embankment is summarized in Table 1. The compaction energy was varied from 50 to 200 blows per unit cell zone using a rapid impact compactor (RIC) to determine the optimal compaction effort for minimizing the construction cost. Three types of soil with different fine contents (18%, 21%, and 33%, respectively) and three embankment heights (0.5, 3, and 5 m) were considered in the study. The 0.5 m height simulated a platform for compacting the existing ground, while the 3 and 5 m heights tested the optimal depth for compaction. Figure 3 shows the layout of the unit cell zones in the field. The compaction energy of 100 blows by the RIC was equivalent to the standard proctor compaction energy.

Table 1 The test program for RIC test embankment

| Parameters | Fill height (m) | | | Fine content (%) | | | No. of blows | | | |
|---|---|---|---|---|---|---|---|---|---|---|
| | 0.5 | 3 | 5 | 18 | 21 | 33 | 50 | 100 | 150 | 200 |

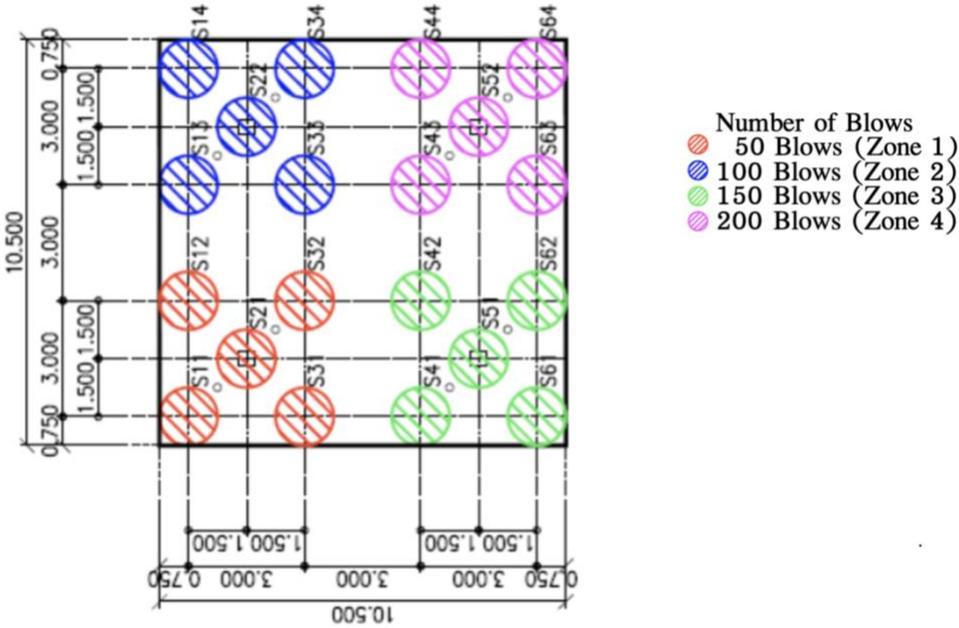

**Figure 3** Compaction layout

### 3.1 Data characteristics
#### 3.1.1 Testing Results
The results suggest that the cone resistance decreases when the number of blows exceeds 100 (Fig. 4). This implies that the optimal compaction effort for RIC is 100 blows per location. Further compaction may cause punching failure and soil weakening. Figure 4 shows the cone penetration test results for different fill thicknesses and compaction efforts. The compaction effect on the underlying soil with high fine content is more pronounced for a fill thickness of 0.5 m, resulting in the lowest increase in cone resistance among the tested cases. The strength of the shallow depth decreases when the compaction effort exceeds 100 blows per location due to punching failure. For fill thicknesses of 3 and 5 m, the cone resistance values of the fill increase significantly with increasing compaction effort until reaching the optimal level of 100 blows per location. Beyond this level, over compaction reduces the cone resistance values. Developing a model that can simulate the behavior of RIC is challenging due to the non-linear relationship between compaction effort and soil strength.

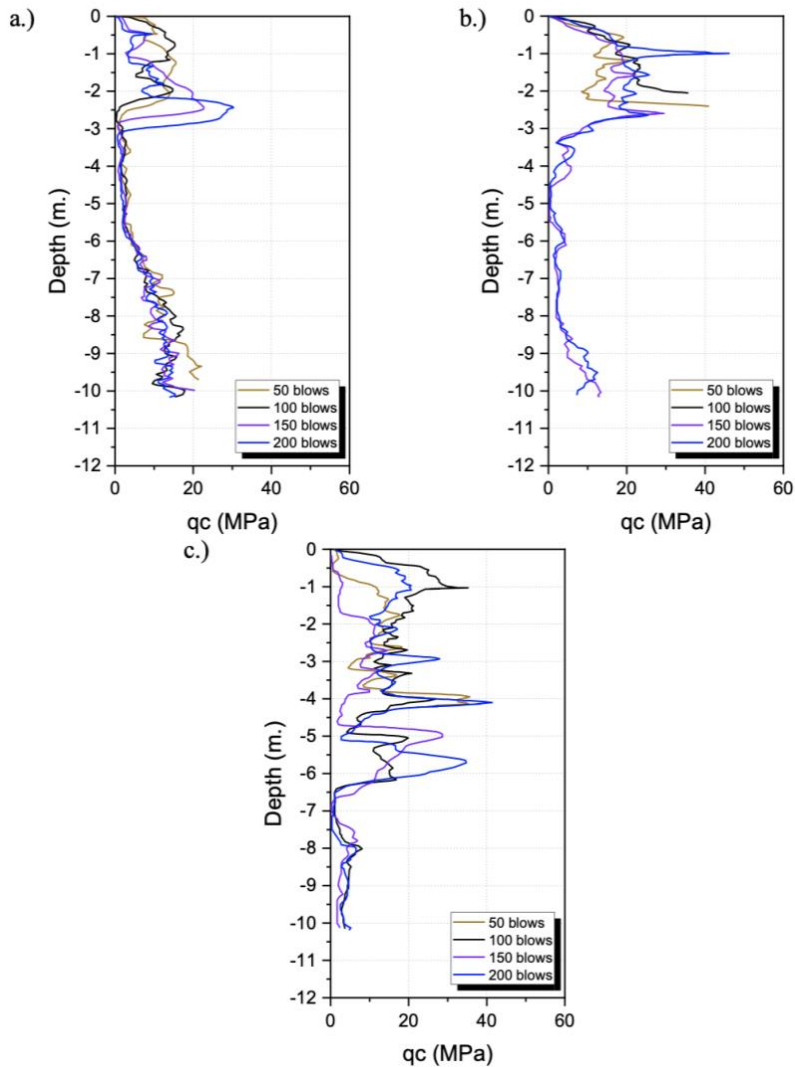

**Figure 4** Cone resistance, qc of soil with fine content at 18 % with fill thickness of a.) 0.5 m. b.) 3 m. and c.) 5 m. for 50, 100,150 and 200 blows

### 3.1.2 Data analysis

This research uses a dataset from a trial test embankment that includes the results of 32 cone penetration tests (CPTs) at different features (Table 2). The dataset contains 32 CPTs before soil improvement and 32 CPTs after soil improvement. The data is derived from the study of Youwai et. al. (Youwai et al. 2023). The dataset also includes various features of compaction schematics, such as number of blows, thickness of fill, and fine content in the field. The original CPT results were filtered to obtain a constant interval of 0.25 m. The sequential data consists of data from the ground surface to a depth of 7 m, with 28 data points for each test. The sequential data after improvement has the same number of data points as the pre-improvement data.

**Table 2** The data characteristics

| Type of data | Dimension (Sample, Sequential data) | Feature dimension |
|---|---|---|

|  |  | Blow | Fill thickness | Fine content |
|---|---|---|---|---|
| Pre improvement $q_c^{ini}$ | (32,28) | 4 | 3 | 3 |
| Post improvement $q_c^{improve}$ | (32,28) | | | |

In this section, we performed feature engineering to select the most suitable features for the model. We used the initial site condition as the input, and the post improvement value as the output. To measure the correlation of each feature with the output, we applied mutual information regression. This is a technique that estimates the mutual information (MI) between a continuous output variable and one or more input variables. Mutual information quantifies how much one variable reduces the uncertainty about another variable. It is zero if the variables are independent, and higher if they are dependent. Mutual information regression can be useful for feature selection, as it can identify the input variables that have the most relevance to the output variable. It can also capture non-linear relationships between variables, unlike correlation-based methods. We use the liberally from Scikit("sklearn.feature_selection.mutual_info_regression" n.d.) in Python to calculate the mutual information of each feature. The formula for estimating the mutual information is shown below:

$$MI(X;Y) = \iint p(x,y) \log\left(\frac{p(x,y)}{P(x)p(y)}\right) dxdy \qquad (1)$$

Figure 5 shows the mutual information (MI) results for each feature affecting the compaction behavior of rapid impact compaction (RIC). The MI values indicate the degree of correlation between the features and the outcome of compaction. The most influential feature is the initial ground condition, measured by the cone resistance value (qc). This is consistent with the rule of thumb that the initial cone resistance is a critical factor for the compaction performance. The second feature that has a significant correlation with the compaction efficiency is the fine content of fill. The results suggest that the fill with low fine content tends to increase the compaction efficiency, while the fill with high fine content may cause punching failure during compaction. The third feature that affects the compaction behavior is the thickness of fill. The MI values show that a thicker fill leads to a higher compaction efficiency. This can be explained by the fact that the existing soil of this site consists of high fine content and some loose layers, which may be penetrated by the RIC plate and result in strength reduction if the fill platform is not high enough. The last feature that influences the compaction behavior is the number of blows. The MI values show that applying more energy does not necessarily improve the compaction efficiency. In fact, excessive compaction effort can also cause punching failure of the existing ground, as observed by the authors in the field compaction. Therefore, an optimal compaction effort should be selected to achieve the desired compaction outcome. Based on the overall MI score, all features were selected to be embedded into a deep learning model for predicting the compaction outcome from RIC.

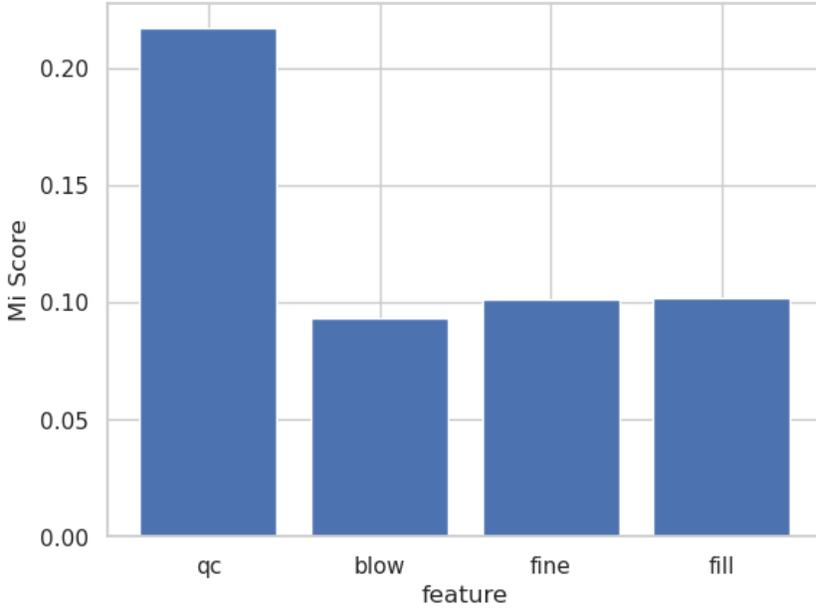

**Figure. 5** The mutual information score of each feature

The efficiency of compaction (EC) to the influence of the feature was shown in this section. The efficiency of compaction is defined as :

$$EC = \left(\frac{q_c^{improve} - q_c^{ini}}{q_c^{ini}}\right) 100 \qquad (2)$$

The effect of different features on the efficiency of compaction (EC) is illustrated in Fig. 6. Generally, the EC was negative when the RIC hammer induced shear failure in the soil layer, resulting in a decrease in the strength of the compacted soil (Fig. 7). This phenomenon poses a challenge for the model simulation, as the parameters of the soil above and below the target level should be incorporated into the model prediction. The bottom row of graphs shows that a higher number of blows improved the strength of the soil below the fill that had a high fine content. The fill with a high fine content might have failed during compaction, causing a reduction in strength and transferring more energy to the lower layer. This phenomenon is similar to the EC with varying thickness of fill. The low thickness of fill with a high fine content also failed during compaction, allowing more compaction effort to reach the lower layer and increase its strength. The top right graph indicates that the lower fine content of fill resulted in a higher EC.

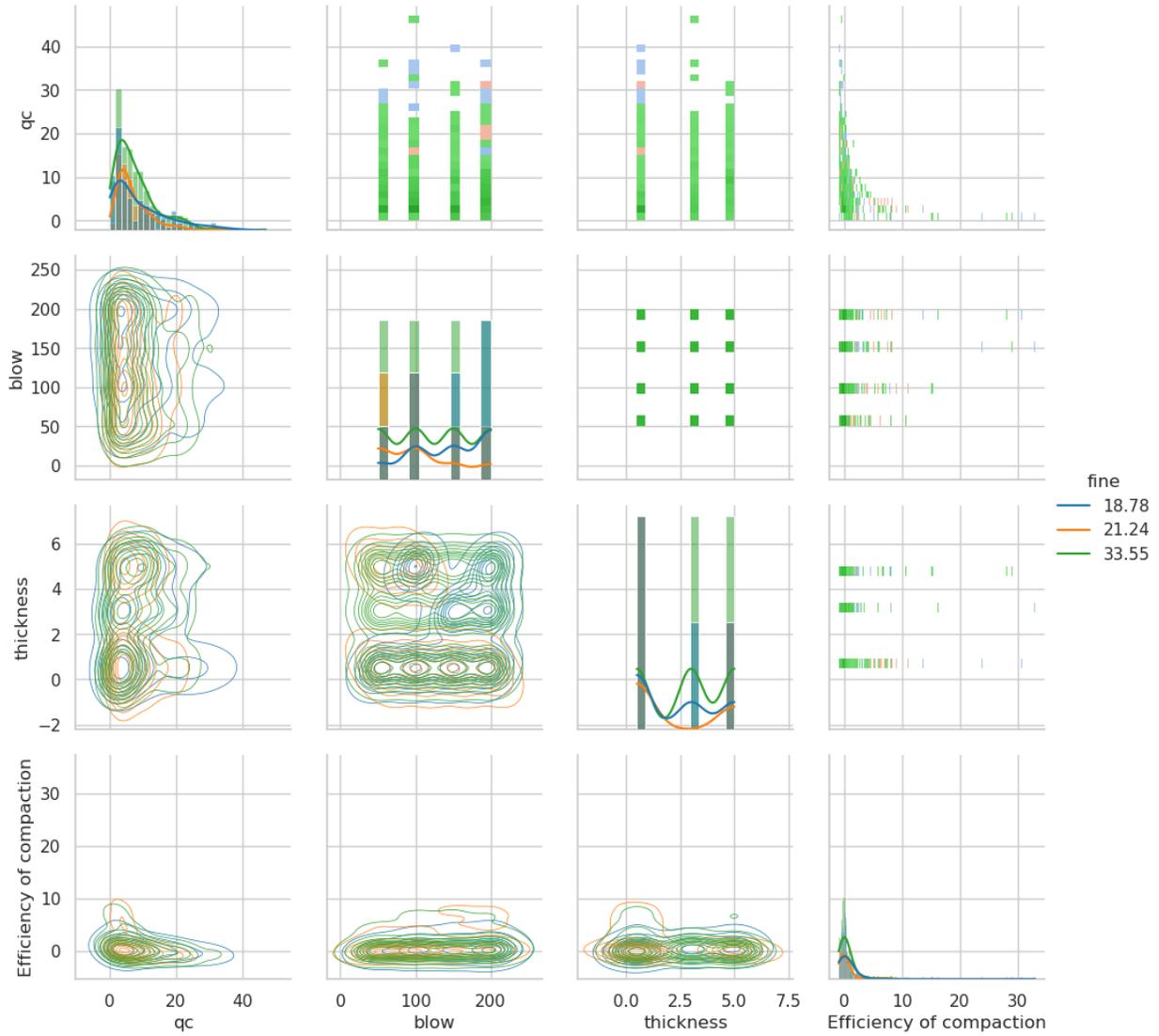

**Figure 6** The overall data characteristics

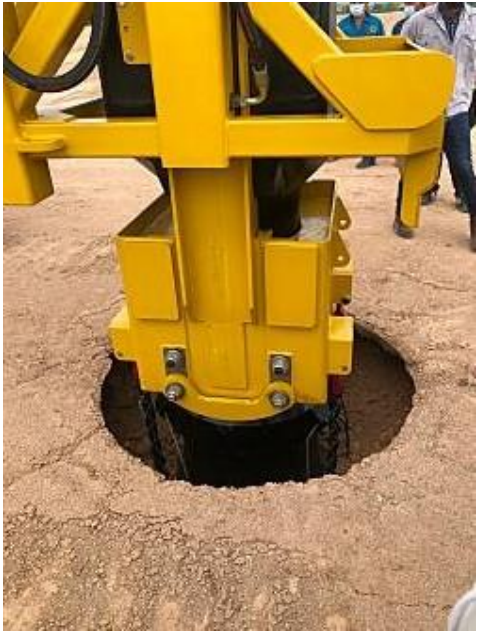 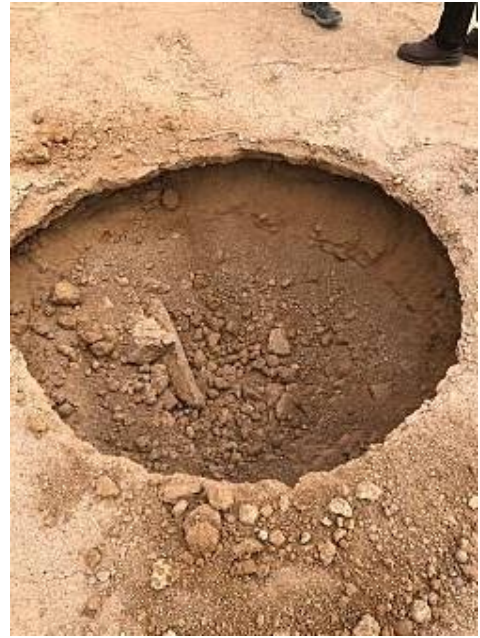

**Figure 7** Punching shar failure of RIC

Kernel density estimation (KDE), a nonparametric method for estimating the probability density function of a random variable, was used to analyze the EC and the thickness of fill, blows, and fine content. KDE does not require any specific assumption about the shape of the underlying distribution, and the density reflects the likelihood of a value in the data. Figs. 8 to 10 show the KDE plots of these variables. The fill thickness did not have a significant effect on EC, as the curves had similar shapes (Fig. 8). The mean of EC was zero, indicating that the compaction was mainly concentrated in the shallow depth (0-4 m). The soil layer below this depth was not influenced by the RIC compaction. The EC ratio was higher for the fill thickness of 0.5 m than for the other fill thicknesses (Fig. 8). However, this fill thickness also had a larger zone of soil strength reduction (negative efficiency of compaction).

The compaction efficiency was significantly influenced by the number of blows and the fine content. The compaction effort showed that the maximum improvement ratio increased with increasing compaction blows until a certain threshold. As shown in Fig. 9, the shape of the kernel density estimation (KDE) was wider for 150 and 200 blows than for lower compaction. This implies that the blows beyond 150 did not result in any benefit in improving the existing ground. Moreover, the soil strength reduction was higher (more than 150 blows) for high compaction effort compared to lower compaction efforts. The reason for this phenomenon is explained in the previous section that excessive compaction created soil failure in the crater of compaction. The soil below the compaction hammer densified to form a platform layer. This caused the soil below the platform layer to exhibit higher improvement ratio. The improvement ratio with different fine content was similar to the previous KDE (Fig. 10). The lowest fine content (18.7%) demonstrated slightly higher improvement ratio compared to other soils with higher fine content due to having a wider shape of the curve. The soil having fine content at 21.24 % showed the lowest effect of compaction compared to other types of soil due to having a narrow shape of the curve and having high density at zero compaction efficiency

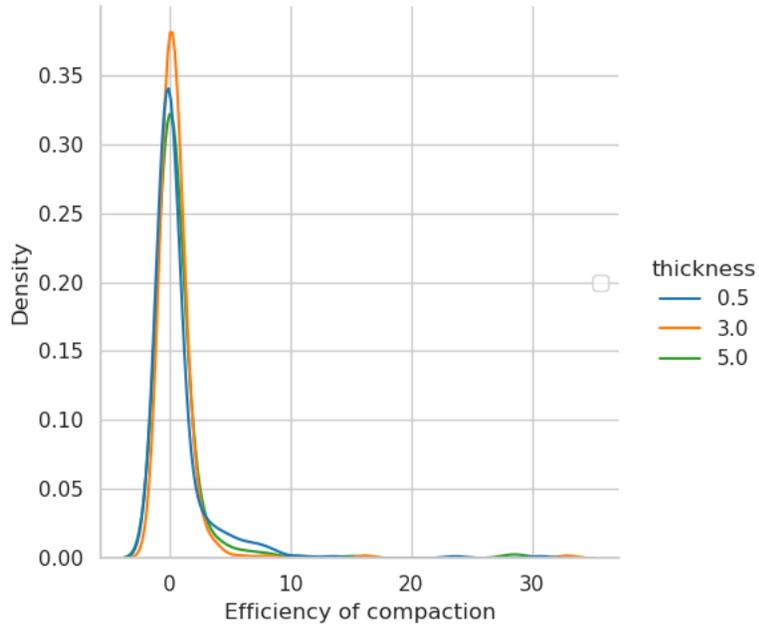

**Figure. 8** Distribution of data for efficiency of compaction with different fill thickness

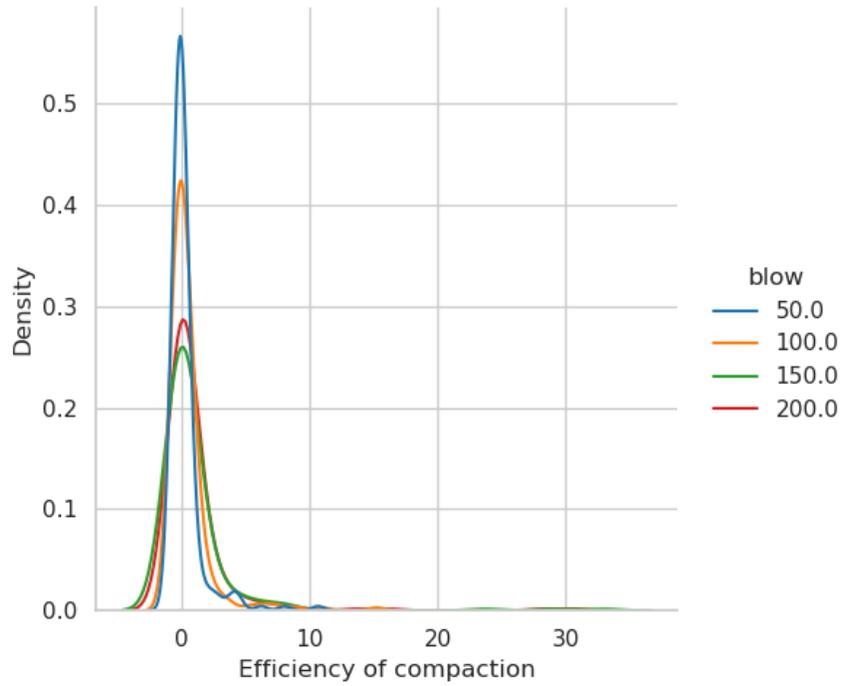

**Figure 9** Distribution of data for efficiency of compaction with different blows

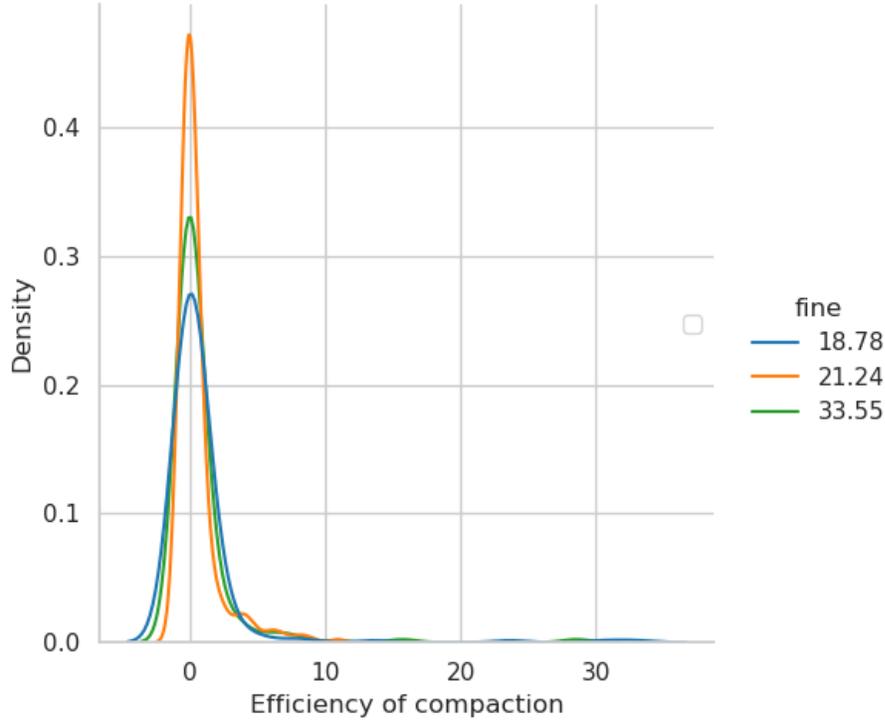

**Figure 10** Distribution of data for efficiency of compaction with different fine content (%)

The feature engineering analysis reveals that the data distribution poses a significant challenge for developing a model to predict the soil strength after RIC improvement. Using the mean value of the improvement ratio as a model parameter is not feasible, as it is approximately zero. Therefore, the model should account for the distribution and variability of each data point in the prediction. The sequence of data is also an important factor for the input value in the established model prediction. In the next chapter, we attempt to develop a deep learning model to predict the RIC behaviors using different architectures. We start with a simple fully connected neural network (Dense), then apply more advanced models that consider the relationship between sequential data, such as LSTM and CNN. Finally, we propose and evaluate a transformer deep learning with hybrid of LSTM and CNN with the same dataset.

## 3. Generative Model

In this study, we developed and evaluated a deep neural network (DNN) model for predicting the outcome for RIC. We used Keras(Chollet and others 2015), a high-level API for TensorFlow(TensorFlow Developers 2023), as the main framework for implementing the DNN model. Keras offers the convenience of using other deep learning libraries, such as PyTorch(Paszke et al. 2019) and Jax(Bradbury et al. 2018), as backends. We trained the DNN model with different architectures and hyperparameters, and compared their performance based on the mean absolute error (MAE) and root mean squared error (RMSE) metrics, defined as follows:

$$RMSE = \sqrt{\frac{\sum(Y_{pred}-Y_{actual})^2}{N}} \qquad (3)$$

$$MAE = \frac{|\sum Y_{pred}-Y_{actual}|}{N} \qquad (4)$$

We split the data into 80% training set, 10% validation set, and 10% test set. The validation set was used to monitor the training process and prevent overfitting. The test set was used to evaluate the generalization ability of the trained model on unseen data. We set the number of epochs to 2000, and used a batch size of 10 and 100 for feed forward and sequence to sequence model, respectively. We used mean squared error (MSE) as the loss function, and adaptive moment distribution method (Adam)(Kingma and Ba 2015) as the optimizer. All of data was used the standard scaler to scale to scale the data before fed it into a neural network. Standard scaler is a technique to transform numerical data by removing the mean and scaling to unit variance. It is useful for machine learning algorithms that assume the input variables have a normal distribution, such as linear regression, logistic regression, and support vector machines. The equation are as follows:

$$X' = \frac{X - \varepsilon}{\sigma} \tag{5}$$

### 3.1 Fully connected neural network (FNN)

This section presents a fully connected neural network (FNN) as a baseline model for comparison with other methods. FNN is a simple neural network architecture that can be applied to regression problems, where the objective is to predict a continuous output variable. This model does not account for the correlation between sequential data, such as the soil properties at different depths in one sample. The prediction task is to minimize the loss value at each specific depth of soil with four features: initial cone resistance ($q_c^{ini}$), blows, thickness of fill (T), and fine content (F). The input data and label are treated as independent at each computation step, as illustrated in Fig. 11. The structure of FNN is summarized in Table 3. The four inputs are fed into a perceptron with weights and biases and passed through a sigmoid(Chen et al. 2019) activation function. This simple model has two layers of perceptron and sends the results to the output layer. The Dropout layer was add into the model to prevent over fitting. The training results yielded a relatively high mean absolute error (MAE) of 0.651 and a root mean square error (RMSE) of 0.9142 (Table 13). These errors suggest that the network architecture did not account for the correlation between the input and output sequences, namely the cone resistance and the relative density index (RIC). The deep learning model attempted to predict the RIC without considering the previous sequential data, which is an important factor for estimating the soil properties. Therefore, to achieve a lower error, the correlation between sequential data should be extracted and embedded into the data before applying the deep learning model.

**Figure 11** Schematic diagram of FNN with single time step

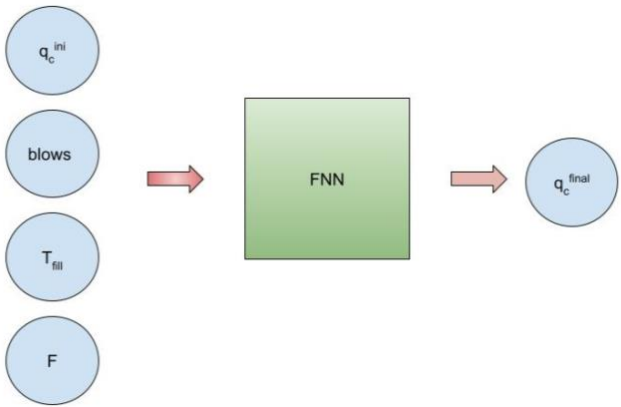

**Table 3** The architecture of fully connected neural (FNN) network

| Layer | No. of perceptron | |
|---|---|---|
| **Input layer** | | **Activation function** |
| Dense | 100 | Sigmoid |
| Dropout | 0.2 | |
| Dense | 50 | Sigmoid |
| **Output** | | |

To incorporate all sequential data into the neural network, the structure of FNN was modified to allow simultaneous input of multiple data points. The first layer of perceptron captured the correlation between sequential data through its weight and bias. Figure 12 illustrates the architecture of the modified FNN_S. The cone resistance data of all soil layers were concatenated with the compaction features, which are blows, fine content, and thickness of fill. The input layer had a dimension of (32,31), where 32 was the number of samples and 31 was the number of features. The output of the deep learning model was the overall outcome of the test results. The prediction scheme is a many-to-many correlation. This model is a generative artificial intelligence approach. The user inputs the initial data and conditions or features that they need, and the model generates the overall output. This concept is analogous to the generative AI for large language models or image generation models, such as Midjourney (Midjourney et al. 2023) and Stable Diffusion (Peng et al. 2018). The geotechnical engineer inputs the soil profile, then the model encodes the soil profile into a latent variable and sends the tensor to the decoding part to produce the desired output. The RMSE of the training dataset decrease significantly from 0.91 to 0.23 (Table 13). However, the RMSE for the testing dataset increased by a small value, which might indicate the overfitting of the trained neural network. This suggests that inputting the overall initial soil condition can improve the performance of the model compared to the case where only the depth that is used for prediction is inputted. The next part will attempt to use a type of neural network that is suitable for sequential data. The over soil profile data will be used as a prime feature to predict the outcome of RIC.

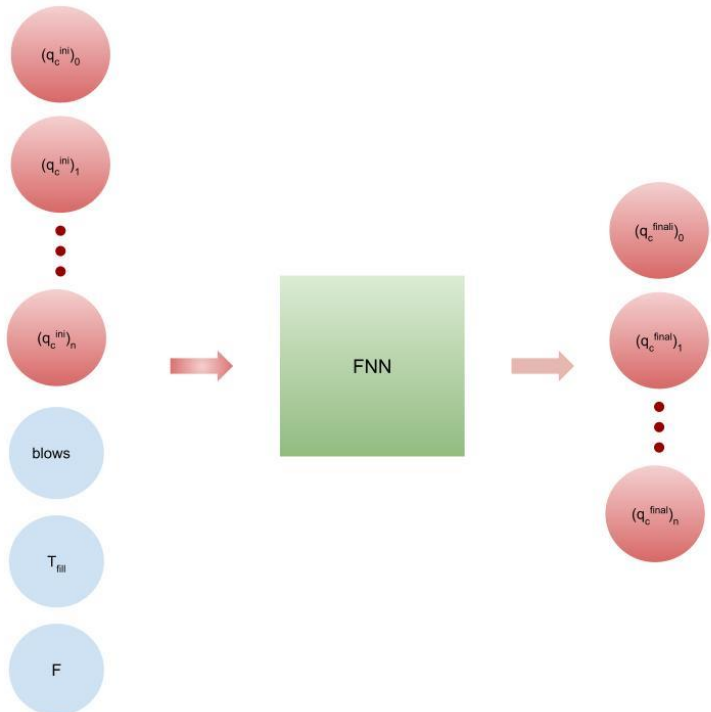

**Figure 12** Schematic diagram of FNN_S with multi time step
**3.2 Sequential deep learning model**

In this section, we applied a sequential model that combines long short-term memory (LSTM) (Hochreiter and Schmidhuber 1997) and convolutional neural network (CNN) (LeCun et al. 1998) to encode the soil profile and feed the tensor to the encoding part for predicting the RIC outcome. The input data was embedded with features into sequential data as shown in Fig. 13. The dimension of the tensor input into the deep learning model was (32,28,4). The first dimension represented the number of training samples. The second dimension was 28 for the number of sequential data points. The last dimension was the number of feature tensors. The first position was the cone resistance result. The second to last layers were the compaction features, which included blows, fill thickness (T), and fine content of fill material (F).

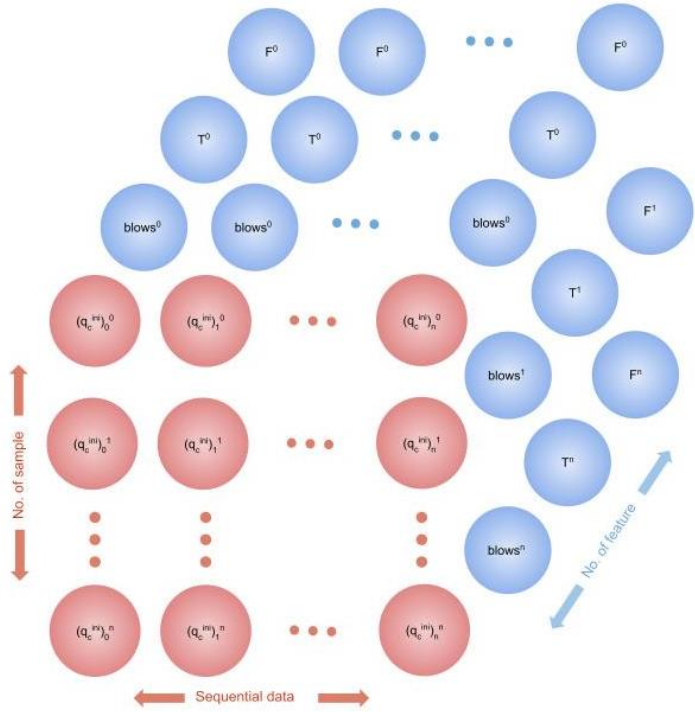

**Figure 13** The input tensor with sequence data and feature was fed into LSTM model.

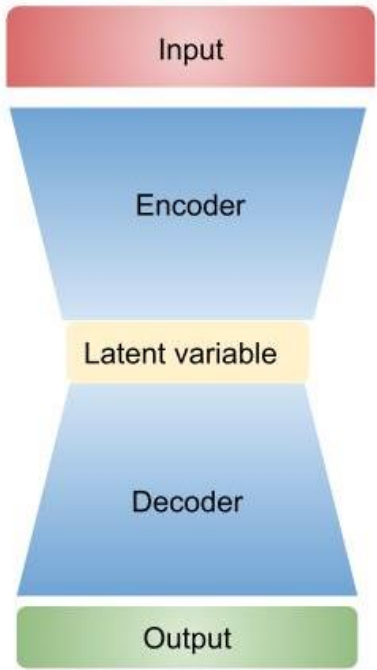

**Figure 14** General architecture of encoder and decoder model

The deep learning model in this study consisted of three parts, as illustrated in Fig. 14. The first part was an encoder, which transformed the soil profile into a tensor using different types of deep learning models and reduced its dimension to a latent variable at the bottleneck layer. The latent variable was then fed to a decoder part, which increased the dimension of the tensor and predicted the output tensor. The back propagation algorithm adjusted the weights and biases of each component to minimize the loss function. The decoder part was fixed for every type of model to ensure the same condition of model comparison.

**3.2.1 Long short-term memory (LSTM)**

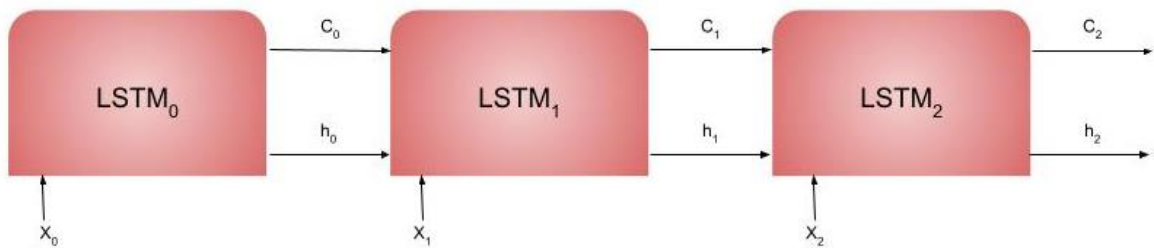

**Figure 15** Architecture of LSTM

Long short-term memory (LSTM) is a type of recurrent neural network that can process sequential data with variable lengths. It has two cell states that store and update information over time, and can learn long-term dependencies between inputs and outputs. Figure 15 shows the architecture of an LSTM network. The input tensor ($x_0$ to $x_n$) is fed into each LSTM cell at each time step. In this study, the input tensor has a dimension of 4, which consists of cone penetration resistance, blows of hammer, thickness of fill and fine content. These values are fed into each cell from the first to the last time step. The hidden state of each cell ($h_n$) represents the short-term memory, and is passed to the next cell and concatenated with the input value. The cell state ($c_n$) represents the long-term memory, and is also passed from one cell to another. The LSTM network can overcome the problem of vanishing or exploding gradients that occurs in standard recurrent neural networks when performing backpropagation. This problem arises when the gradients become too large or too small after many multiplications, making it difficult to optimize the network parameters. The LSTM network uses a mechanism called gates to control the flow of information between cells, and to prevent unwanted changes in the cell states. This way, the LSTM network can achieve a low loss value for sequential data modeling. In this study, every hidden state of cell was combined with final output and fed to the next deep learning layer.

**Table 4** The architecture of input data for LSTM

| Layers | No. of perceptron |
|---|---|
|  | **Normal** LSTM |
| LSTM | 200 |
| Dropout | 0.2 |
| Dense | 200 |
| Dense | 50 |
| Output-latent space | |

**Table 5** The architecture of decoder

| Layer | No. of perceptron |
|---|---|
| Input -latent space | |
| Dense | 200 |
| Dense | 100 |
| Output | |

The input tensor, as shown in Table 4, was fed into an LSTM layer with 200 units to process the sequential data. The LSTM layer is a type of recurrent neural network that can learn long-term dependencies between inputs and outputs, and has two cell states that store and update information over time. To avoid overfitting of the model, a dropout layer with a rate of 20 % was applied as a regularization technique. The output from the LSTM layer was then passed to two feed-forward neural network layers with 200 and 50 units, respectively (Table 5). The output of the last layer, with a dimension of 50, was used as a latent variable for the decoding layer, as shown in Table 5. The decoding layer consisted of two feed-forward neural network layers with 200 and 100 units, respectively, followed by an output layer that generated the predicted values. The output of this study is the value of cone resistance at every soil layer with 0.25 m spacing for 7 m depth. The dimension of the output is 28. The LSTM model achieved a significant improvement in MAE and RMSE, reducing them by about four times compared to the FNN model (Table 11, Fig 20). However, the test error did not show a significant improvement, indicating a high degree of overfitting for the LSTM model. Moreover, the LSTM model had a large number of trainable parameters (487,810), which increased the training time per epoch by three times compared to the FNN model. The MAPE obtained by the LSTM model was still unsatisfactory for the authors. Therefore, another type of sequential model, convolutional neural network (CNN), was assessed in the next part.

**3.2.2 One-Dimentional Convolutional Neural network (CNN)**

One-dimensional convolutional neural networks (1D CNNs) are a type of deep learning model that can process one-dimensional data, such as sound signals or time series, and perform tasks such as classification or regression (Fig. 16). A 1D CNN consists of convolutional, pooling, and fully connected layers. The convolutional layers apply a one-dimensional filter to the input data, extracting local features. The pooling layers reduce the dimensionality and increase the invariance of the feature maps. The fully connected layers use the learned features for the final task. 1D CNNs have several pros and cons for applications that involve sequential or temporal data. They can learn from the raw data without manual feature engineering, capture long-term dependencies and patterns, and handle variable-length inputs. However, they can also be computationally expensive, require a large amount of training data, suffer from overfitting and underfitting, and be sensitive to noise and outliers.

To encode the input data into a 50-dimensional latent variable, we employed a one-dimensional convolutional neural network (CNN) model (Table 6). The model comprised two convolutional layers with 128 and 64 filters of size 4 and 2, respectively, followed by a flattening layer. The flattened output was then decoded by a structure similar to the LSTM model (Table 5).

The results of RMSE and MAE (Table 13) showed that the LSTM model outperformed the CNN model slightly. This may be attributed to the ability of LSTM to capture long-term dependencies and temporal patterns in sequential data better than CNN. LSTM has a memory cell that can store and update information over time, while CNN relies on convolutional filters and pooling layers to extract features from local regions of the input. LSTM may be more suitable for tasks that require long-term context and complex temporal dynamics. However, CNN demonstrated slightly better generalization to the test data, indicating lower overfitting (Table 11, Fig 20). The CNN model can reduce the dimensionality of the input data and prevent overfitting by using convolutional filters and pooling layers. Nevertheless, both sequential models, LSTM and CNN, have a high error and overfitting. We will continue to develop deep learning models in the next part.

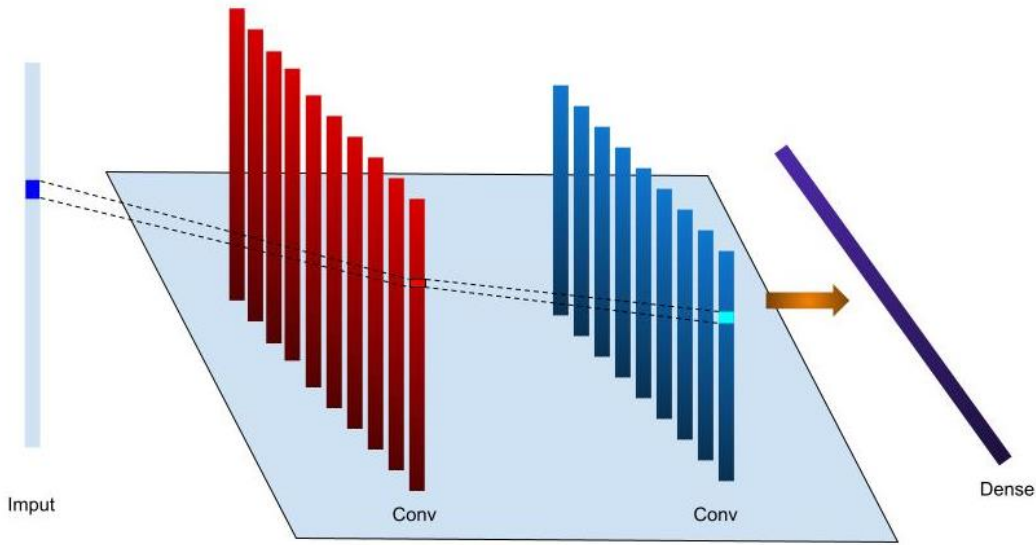

**Figure 16** The architecture of convolutional network model

**Table 6** The structure of CNN model

| Layer | No. of perceptron |
|---|---|
| | **CNN** |
| Con1D (filter/kernel_size) | 128/4 |
| Con1D (filter/kernel_size) | 64/2 |
| Flatten | |
| Dropout | 0.2 |
| Dense | 50 |
| **Output-latent space** | |

### 3.2.3 Hybrid LSTM-CNN Model

To reduce the prediction error of the outcome of RIC, we combined LSTM and CNN in a hybrid model (Fig. 17). A hybrid LSTM_CNN model is a neural network that leverages the strengths of both long short-term memory (LSTM) and convolutional neural network (CNN). It can process sequential and spatial data, such as text, images, audio, or video, and extract both local and global features of the data. Compared to a single model, it can handle more complex and diverse data types, achieve higher accuracy, and reduce the number of parameters and computational cost. The hybrid model has the same input tensor as the previous models (LSTM and CNN). The structure of the two sequential models is similar to the previous section (Table 4 and Table 6). The outputs of both models are concatenated together to form a latent variable. The latent variable is then fed into the decoding model (Table 5), which decodes it and produces a prediction of the output. To further reduce the error, we added a condition tensor (feature) to the latent variable, following the same approach as the conditional variational autoencoder (CVAE) (Fang et al. 2021). The feature input is passed through a fully connected neural network with a dimension of 40 (Table 7). We added this layer to introduce trainable weights for the feature before concatenating it with the latent variable. The latent variable is then concatenated with the tensor from the feature input. Finally, the latent variable is used to decode the tensor to the desired output.

The hybrid model, which combined the strengths and reduced the weaknesses of each single model, showed a slightly better performance than the LSTM and CNN models in terms of RMSE and MAE (Table 11). The test data set error of the hybrid model was about 14% lower than that of the single models. However, the many-to-many deep learning (MDL) models still produced overfitting behaviors. A possible reason for the failure of the MDLs to simulate the outcome of RIC was that they did not consider the previous predicted sequence value in the decoding model. The soil properties in the deeper layer could depend on the soil properties in the upper layer, which could be explained by the energy transfer due to the compaction. If the soil in the shallow layer had a lot of improvement, it meant that it absorbed high energy from the compaction, resulting in low energy transfer to the deeper layer. In the next part, we adopted the concept of sequence to sequence in the transformer deep learning for large language model. The decoder of the language model attempted to predict the next word of sentence by using the encoding tensor from the previous time step, the word that it predicted from the previous step, combined with latent variable from the decoding part. We expected that this architecture might be useful for a model to predict the outcome from RIC.

**Table 7** The structure features encoder

| Layer | No. of perceptron |
|---|---|
| Input | |
| **Dense** | 40 |
| Output-Latent space | |

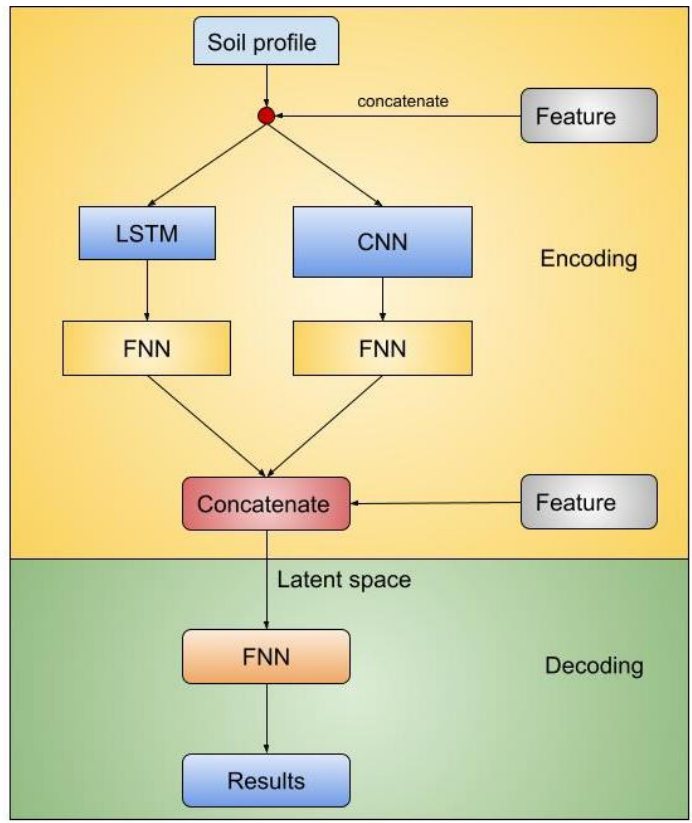

**Figure 17** Architecture of hybrid LSTM-CNN model

**3.3 Sequence to sequence transformer model**

This part aims to simulate the outcome of RIC by using a sequence-to-sequence model. The structure of the model is based on the original transformer model (Vaswani et al. 2017). The transformer model was adapted to handle a regression problem. The next part explores the application of recurrent neural network and CNN to the sequence-to-sequence model with different architectures. The details of each model will be presented in the following section.

**3.3.1 Transformer Model**

In this part, we aim to develop a sequence-to-sequence model to predict the outcome of RIC. The architecture is based on the paper 'Attention Is All You Need' proposed by (Vaswani et al. 2017). The original paper developed an attention-based model for the task of neural machine translation. The schematic diagram of the transformer model is shown in Fig. 18. This paper sparked a revolution in natural language processing, leading to the development of large-scale language models such as GPT (OpenAI 2023), BERT, and LLAma (Touvron et al. 2023). These language models are essentially classification problems, as they try to predict the next word or the class that provides the highest probability. However, the research on using transformer models for regression problems is limited (Born and Manica 2022, 2023; Su et al. 2022). Therefore, adapting the transformer model for regression tasks poses a significant challenge. The original transformer model contains several components that use activation functions such as SoftMax, which are suitable for classification tasks. The weights of each neuron are normalized to sum up to one or unity. This could be problematic for regression tasks that require actual values rather than probabilities to select the correct category or class in language models.

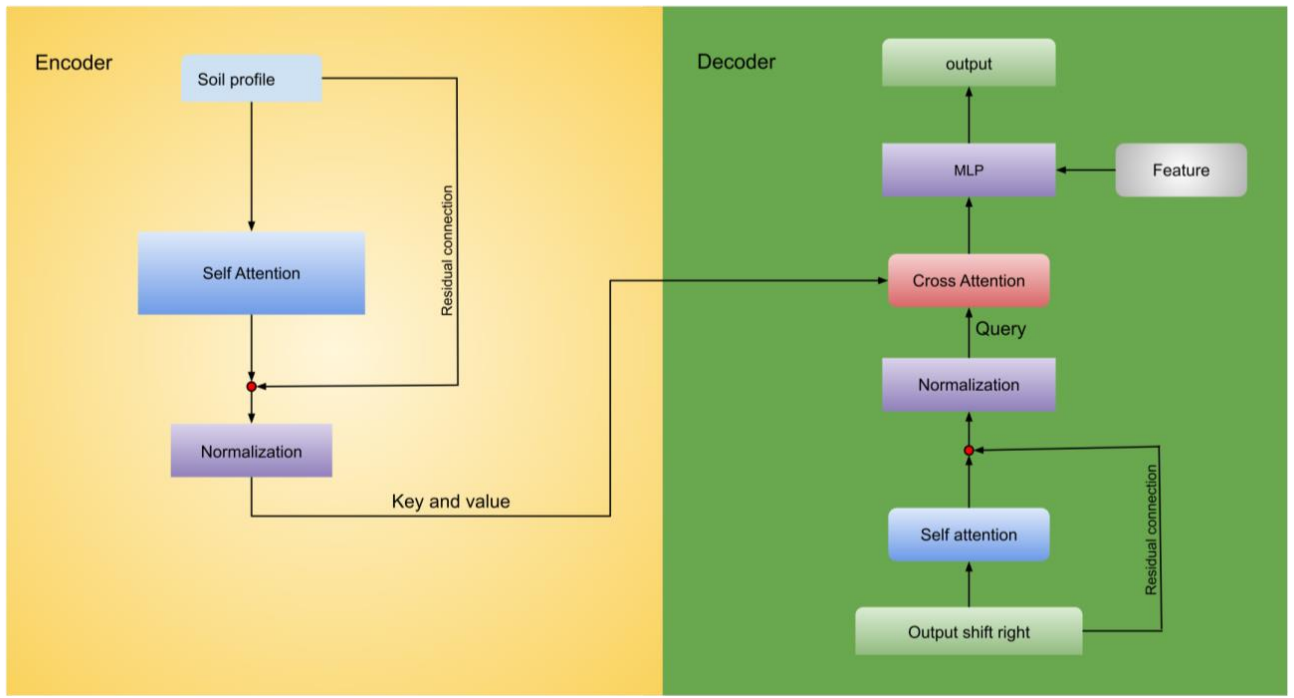

Figure 18 The architecture of transformer model

Table 8 The architecture of transformer model detail

| Encoder | Self attention: head =64, num_head=8 | 1 layers |
|---|---|---|
|  | Normalization |  |
| Decoder | Self attention: head =64, num_head=8, dropout=0.1 | 1 layers |
|  | Normalization |  |
|  | Cross attention: head =64, num_head=8, dropout=0.1 | with encoder |
|  | Dense (50) | Sigmoid |
|  | Dense (28) | Sigmoid |
|  | Concatenate -Feature vector |  |
|  | Dropout 0.1 |  |
|  | Multi level perceptron Dense (100)-(50)-(10)-(5) | Sigmoid |
|  | Sigmoid | Linear |

The model architecture consists of two components: an encoder and a decoder, as shown in Fig. 18 and Table 8. The model extracts core features based on the concept of attention, which is a technique that enables a neural network to learn the relationships between different parts of a sequence, such as words in a sentence or pixels in an image. It computes a score for each pair of elements in the sequence, and then uses these scores to create a weighted representation of the sequence. Self-attention can be used to encode, decode, or generate sequences, and it has many applications in natural language processing, computer vision, and generative models. One of the advantages of self-attention is that it can capture long-range dependencies in the data, unlike recurrent or convolutional networks that rely on local information. Another advantage is that it can be parallelized easily, which makes it faster and more scalable. A popular model that uses self-attention is the Transformer (Vaswani et al. 2017) which has achieved state-of-the-art results in many tasks such as machine translation, text summarization and image captioning (Zohourianshahzadi and Kalita 2021). The attention can be simply as equation as follows:

$$Attention(Q,K,V) = softmax\left(\frac{QK^T}{\sqrt{d_k}}\right)V \qquad (6)$$

The model employs multi-head attention (Table 8) to extract features from the input soil profile, following the original paper 'Attention Is All You Need' (Vaswani et al. 2017). The model has a head size of 64 and a number of heads of 8. Unlike the word embedding for language models, which uses two-dimensional input vectors, this study adopts one-dimensional input vectors to simplify the model and reduce the training time. The attention layer is set to one layer based on hyperparameter tuning, which yields a lower loss than higher layers of attention. Higher layers of attention may introduce more non-linearities and dependencies among the inputs and outputs, making the optimization landscape more challenging and prone to local minima (Wang et al. 2021b). The RIC feature is incorporated into the model in the encoding part, after the attention output is added to its initial value as a residual technique to prevent gradient vanishing when running with many loops. Then, the result is normalized and sent to the decoder part. For the decoder part, the output of the previous step (output shifted right) is the input vector of encoding. Then, the vector is sent to the self-attention layer to extract features from the input vector. Then, the vector is normalized and sent to the cross-attention layer with the vector from the encoder part. Then, the results from cross-attention are concatenated with the vector that represents the RIC features such as compaction effort, depth of compaction platform, and fine content of soil.

### 3.3.2 Hybrid CNN and LSTM sequence to sequence model

A sequence-to-sequence model is a type of generative model that consists of two components: an encoder and a decoder (Fig. 19). The encoder is composed of a sequential model, such as a recurrent neural network (RNN) or a transformer, that processes the input sequence and encodes it into a fixed-length vector. The encoder also has a multilayer perceptron (MLP) that reduces the dimension of the vector and sends it to the decoder as a latent variable. A latent variable is a hidden representation that captures the essential information of the input sequence. The decoder is another sequential model that receives the latent variable and the previous output sequence (output shifted right) as inputs, and generates the next output sequence. The decoder also has an MLP that concatenates the output vector from the sequential model with the latent variable, and then produces the desired output. A sequence to sequence model can be used for various tasks such as machine translation, text summarization, speech recognition, and image captioning.

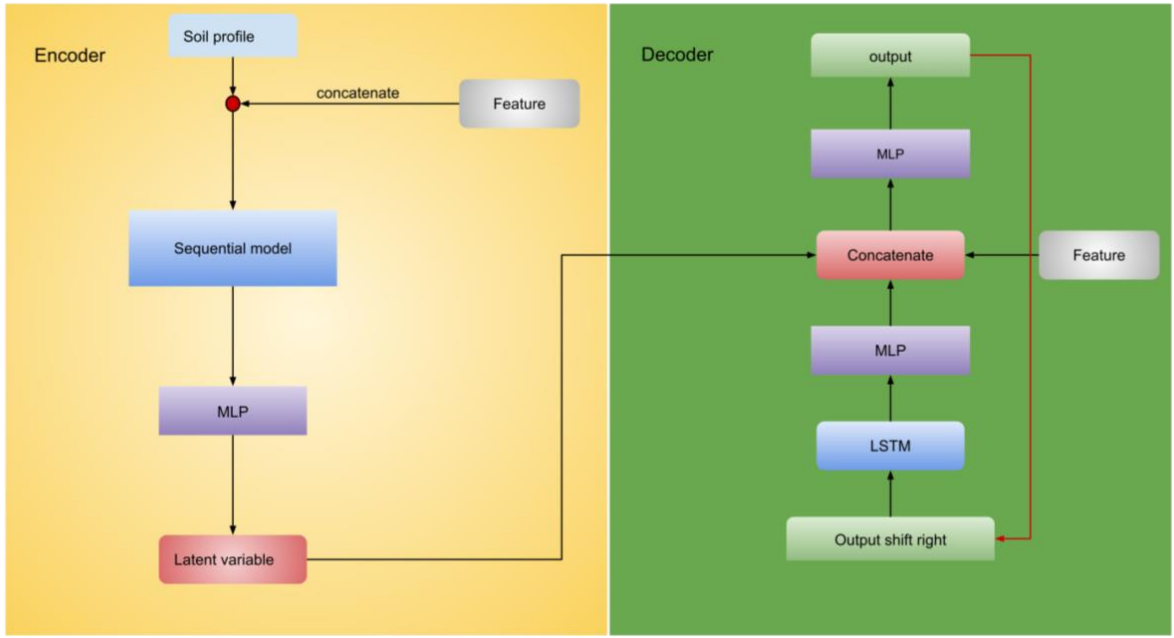

**Figure 19** The architecture of sequence-to-sequence model

Incorporating features into the calculation is a challenge for modeling the outcome of RIC. The RIC consists of different compaction efforts, fine contents, and thicknesses of fill. The features are added to the model in both the encoder and decoder parts. The features are appended to the input tensor before feeding it into a sequential model (Fig. 19). The concept is similar to adding features for the sequential model in the previous part (Fig. 17). The features and the sequential data are passed through the feature extraction by RNN and CNN models. Then their dimensions are reduced to the desired dimension of the latent variable. The features are also added to the model again in the decoding part. They are concatenated with the results from the previous output sequential data and the latent variable. The two-time addition of features or conditions to the model is similar to the architecture of Conditional Generative Adversarial Network (cGAN) (Mirza and Osindero 2014). The condition is added in both locations before the generator and before the discriminator.

In this study, we attempted to compared different of sequential model in the encoder with single model LSTM (LSTM-S2S), CNN ((LSTM-S2S) and hybrid model between LSTM and CNN (LSTM_CNN-S2S). The structure of the CNN and LSTM was similar to the previous part for good in comparison (Table 9). The decoder takes three input tensors as shown in Fig. 19 and Table 10: the latent variable, the feature tensor, and the output shift right. The latent variable is derived from the encoder that captures the initial soil condition and features. The feature tensor has a dimension of 16 and is obtained by applying a feed-forward neural network (FFN) layer with a dimension of 40 to the input feature tensor (Table 9). The output shift right is a technique (Vaswani et al. 2017) used in the Transformer model to prepare the input and output sequences for the decoder. It involves adding a special token 0 at the beginning of the output sequence and shifting the rest of the tokens one position to the right. This way, the decoder can predict the next token based on the previous tokens, without peeking at the current token. The sequence token is padded with zeros for the missing values. The output shift right is then passed through a long short-term memory (LSTM) layer and then an MLP layer to reduce the tensor dimension to 16 (Table 9). The concatenated tensor is then fed into a dropout layer with a rate of 0.2 to reduce overfitting. The tensor is then fed into another MLP layer to reduce the dimension from 30 to 5 and then to one for output value.

**Table 9** Encoding and decoding model architecture

| Encoder | LSTM Model | CNN Model |
|---|---|---|
| | LSTM 200 | Con1D (128/4) |
| | Dropout 0.2 | Con1D (64/2) |
| | Dense 200 | Dropout 0.2 |
| | Dense 50 | Dense 50 |
| | Latent variable (16) | |
| Decoder | Output shift right | Feature |
| | LSTM (100) | Input (None,3) |
| | MLP-Dense (40,16) | MLP-Dense (40,16) |
| | Concatenate | Latent variable<br>Output shift right<br>Feature |
| | Dropout 0.2 | |
| | MLP -Dense (30, 5, 1) | |
| | Output(1) | |

To reduce the error of the deep learning model, we incorporated a self-attention mechanism into the LSTM-ATT-S2S model. We modified the CNN branch of the model by adding a self-attention layer before feeding the vector to the CNN layer. We also adopted the multi-head attention mechanism for the CNN component of our model, which we named the transformer, as shown in Table 10 and Fig. 20. The model followed the procedure described by (Vaswani et al. 2017), for the attention process. First, the model applied layer normalization to the input tensor, which is a technique that normalizes the input or output of each sub-layer in the transformer architecture. Layer normalization reduces the feature variance and enhances the stability and performance of the neural network training. Next, the model computed a multi-head self-attention layer on the normalized tensor, which calculated a weighted sum of the input features according to their relevance to each other. The output of the self-attention layer was then added to the input tensor via a residual connection, which created a shortcut path for the gradient to propagate back during backpropagation. Residual connections also help to preserve some information from the original input throughout the network, which can improve the representation learning. The resulting tensor was then normalized again using another layer normalization technique.

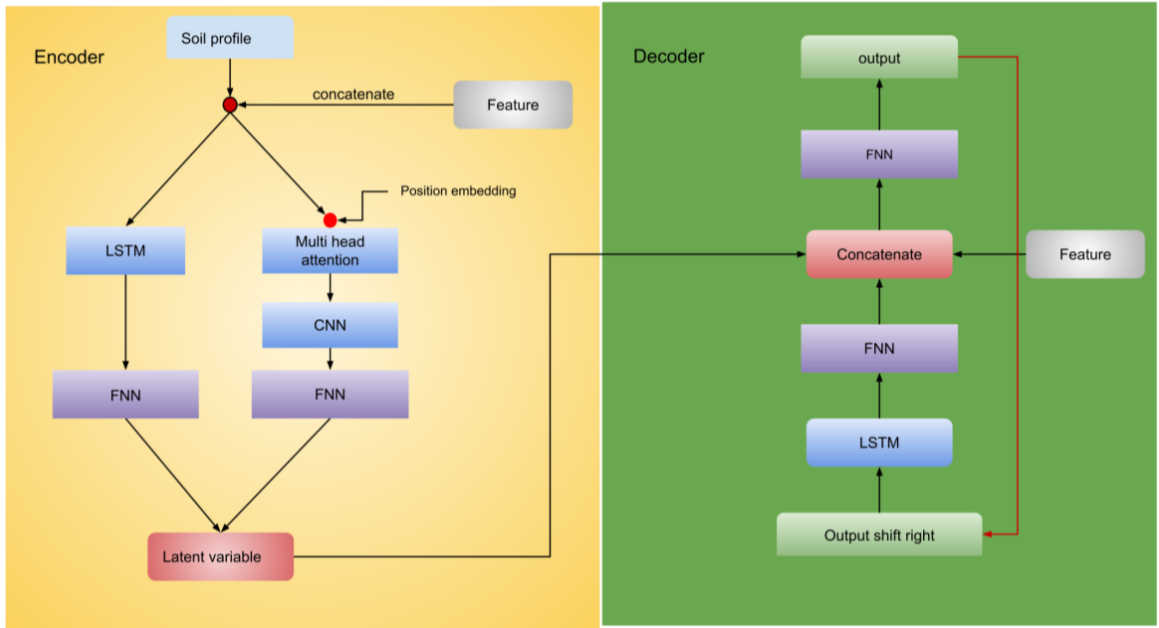

**Figure 20** Architecture of transformer model

The embedded position encoding was also integrated to the model to embedded the position of certain soil layer to training with embedded vector. Embedded position encoding is a technique to represent the position of each token in a sequence as a vector. It is used in transformer models, which do not have recurrence or convolution, to capture the order of the tokens. Embedded position encoding is learned during the training process, unlike positional encoding, which is fixed and based on mathematical functions. Embedded position encoding can adapt to different tasks and domains, and may capture more complex patterns than positional encoding. We use the liberally provide for using in natural language processing (NLP) in Keras (Devlin et al. 2019; Team n.d.) The equation of position embedded is as follows:

$$PE(pos, 2i) = \sin\left(\frac{pos}{100^{\frac{2i}{d_{model}}}}\right) \quad (7)$$

$$PE(pos, 2i+1) = \cos\left(\frac{pos}{100^{\frac{2i}{d_{model}}}}\right) \quad (8)$$

After that, the tensor underwent two layers of CNN for feature extraction (Table 10). Finally, we added another residual connection to the output tensor of the CNN layers. The tensor from transformer part and LSTM (Table 10) was concatenate together and yield the latent variable. It will send to decoder model section for estimate the cone bearing after improvement. During calculation the encoding model perform only one time and send the results to generate in the decoding part of model.

**Table 10** Model detail and parameter of transformer.

| Layer | Model detail |
|---|---|
| Input | |
| Position embedding | |
| Layer Normalization | $\varepsilon = 10\text{-}6$ |
| Multi Head Attention | head_size=24, num_heads=10 |
| Residual connection | |
| Layer Normalization | $\varepsilon = 10\text{-}6$ |
| Con1D (filter/kernel_size) | (2,1) |
| Con1D (filter/kernel_size) | (4,1) |
| Residual connection | |
| Flatten | |
| Dense | 50 |
| Latent variable | |

## 3.4 Results and Discussion

A comparison of the accuracy and training time of the sequence to sequence (Seq2Seq) model and the feed-forward model was conducted and the results are shown in Table 11 and Fig. 20. The Seq2Seq model achieved lower errors for both test and train data than the feed-forward model, which exhibited overfitting when the model complexity increased. This can be explained by the fact that the Seq2Seq model incorporated the previous output prediction into its architecture, while the feed-forward model only used the input data, initial soil profile and feature, to predict an output. Although the LSTM and CNN models had lower RMSE values than the Seq2Seq model, they also had higher RMSE values for testing data, indicating a trade-off between accuracy and generalization. The complexity of the Seq2Seq model seemed to have a positive effect on the prediction accuracy, but it also increased the training time by about 300 times compared to the base model (FNN).

The hybrid Seq2Seq model outperforms the original transformer model in terms of error rates on both the training and test data sets, as shown in Fig. 20. This might be attributed to the structure of the transformer model, which normalizes the tensor multiple times using the SoftMax function and a normalization layer before adding the residual. The transformer model is also sensitive to the choice of hyperparameters, which makes it difficult to train, as reported by many researchers(Liu et al. 2020). Increasing the complexity of the model can lead to instability or suboptimal performance, resulting in a zero gradient. However, the transformer model has an advantage in its low training time per epoch, which is the lowest among the Seq2Seq models. The low training time is due to its simplified input vector, which has only two dimensions instead of three, and its parallel computation, which allows it to process each input token simultaneously, unlike the sequential models that have to wait for the previous token's calculation.

Soil compaction prediction can be improved by using a more complex model that incorporates LSTM, CNN, and attention mechanisms (Fig. 21 and 22). The proposed LSTM-ATT-S2S model extracts the most relevant features from the input sequence and feeds them into the decoder to generate the target value at each time step. This model achieves the lowest prediction error among the compared models. Transformer model has a higher prediction error when the actual data shows abrupt changes in value, as seen in Fig. 22b at depth 1.5 to 2.5. This is also reflected in the prediction of different compaction features in Fig. 22. The Seq2Seq model uses the previous actual output, shifted right by one time step, as the input for the current prediction. This shows that the Seq2Seq model can fit the data well, and the LSTM-ATT-S2S model performs slightly better than other models. However, for generative models, high accuracy is required because the previous prediction is used as the input for the next step, which could lead to cumulative errors when generating long sequences of tokens. The last token in the sequence will contain a lot of errors. On the other hand, the LSTM-ATT-S2S model takes much longer time to train and compute than other simpler models. Therefore, the appropriate model architecture should be selected according to the purpose and application of the model, balancing the trade-off between accuracy and efficiency. The training process might also require a powerful computer and create a lot of carbon footprint.

**Table 11** The comparison with different generative model

| Model | MAE | | RMSE | | Trained Parameters | Training Time/epoch ms |
|---|---|---|---|---|---|---|
| | Train | Test | Train | Test | | |
| Feed forward model | | | | | | |
| FNN (base model) | 0.652 | 0.6235 | 0.9142 | 0.8734 | 301 | 12 |
| FNN_S | 0.1680 | 0.5417 | 0.2311 | 0.7904 | 9,378 | 26 |
| LSTM | 0.0392 | 0.6117 | 0.0538 | 0.9109 | 492,412 | 63 |
| CNN | 0.0437 | 0.5916 | 0.0574 | 0.9081 | 139,234 | 52 |
| LSTM_CNN | 0.0394 | 0.5112 | 0.0519 | 0.7906 | 593,284 | 67 |
| Sequence to sequence model | | | | | | |
| Transformer | 0.1292 | 0.5020 | 0.1868 | 0.7494 | 192,399 | 200 |
| LSTM-S2S | 0.0947 | 0.4431 | 0.1425 | 0.6486 | 448,435 | 840 |
| CNN-S2S | 0.0934 | 0.4490 | 0.1439 | 0.6553 | 256,385 | 479 |
| LSTM_CNN-S2S | 0.0989 | 0.4368 | 0.1572 | 0.6525 | 448,435 | 905 |
| LSTM-ATT-S2S | 0.0982 | 0.4264 | 0.1446 | 0.6303 | 866,573 | 3,000 |

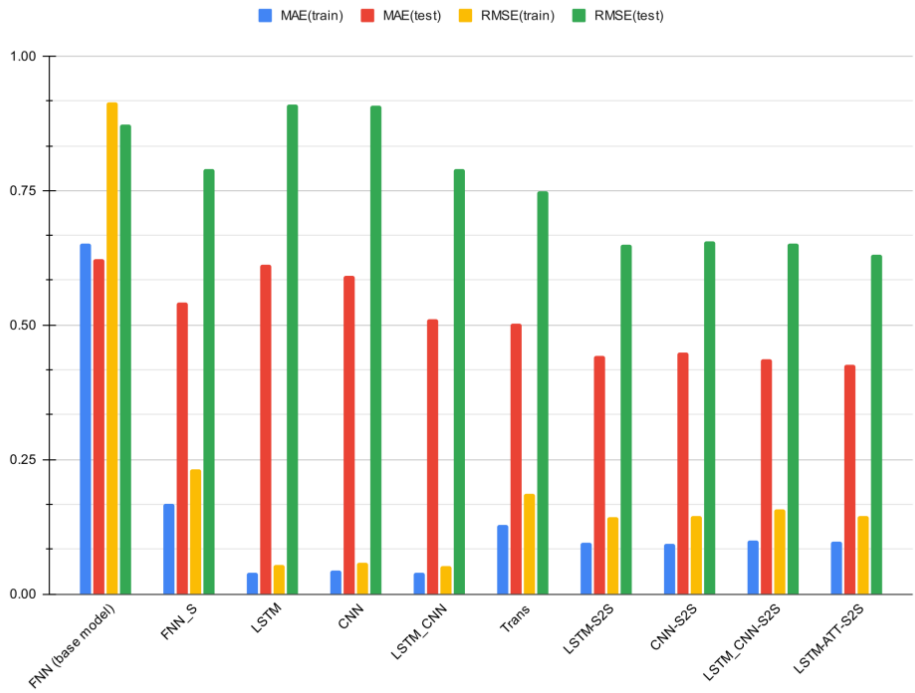

**Figure 20** Comparison of the MAE and RSME of each deep learning model

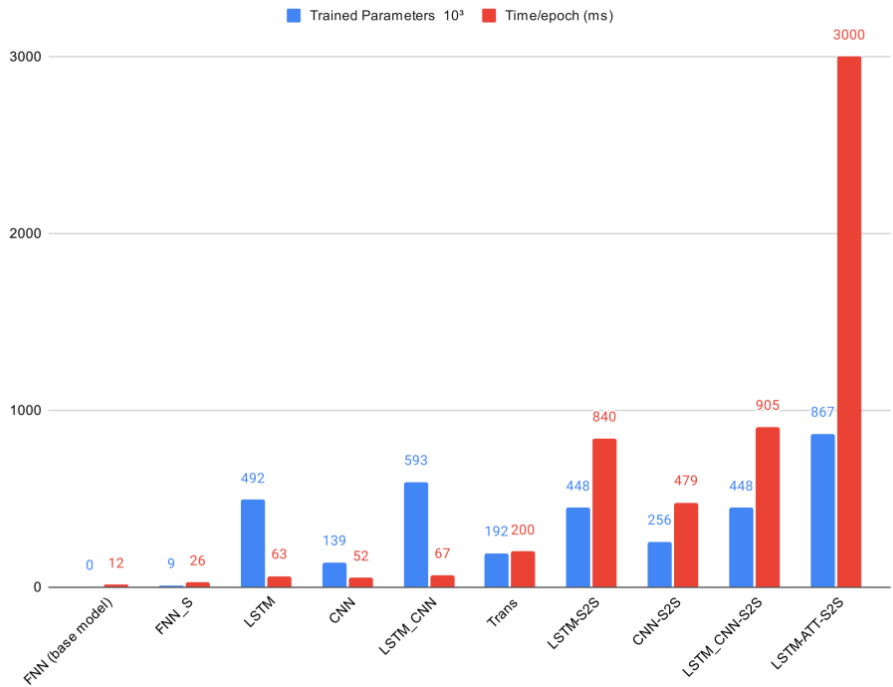

**Figure 21** Trained parameters and time for training per epoch of different models

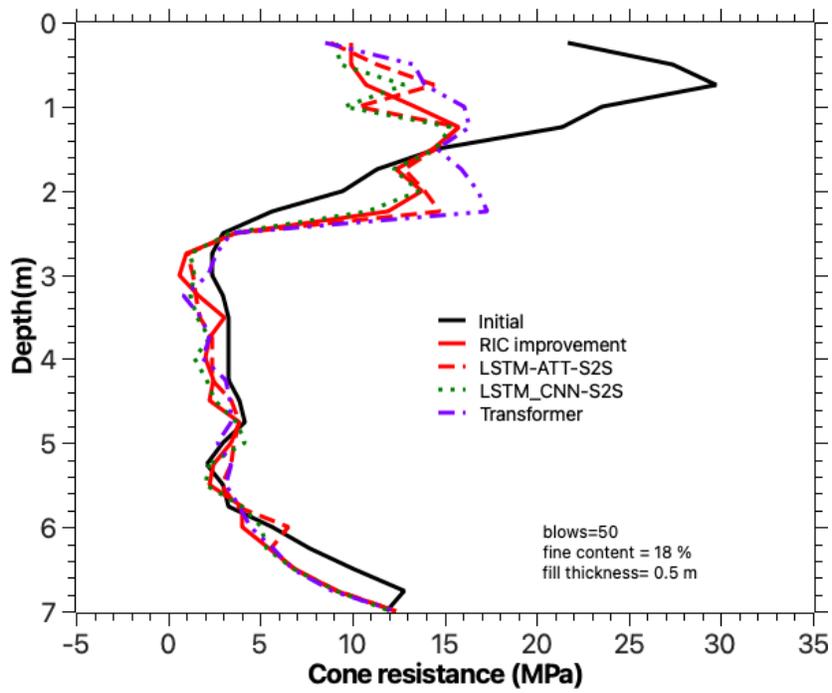
a) blows =50, fine content =18 % and fill thickness=0.5 m

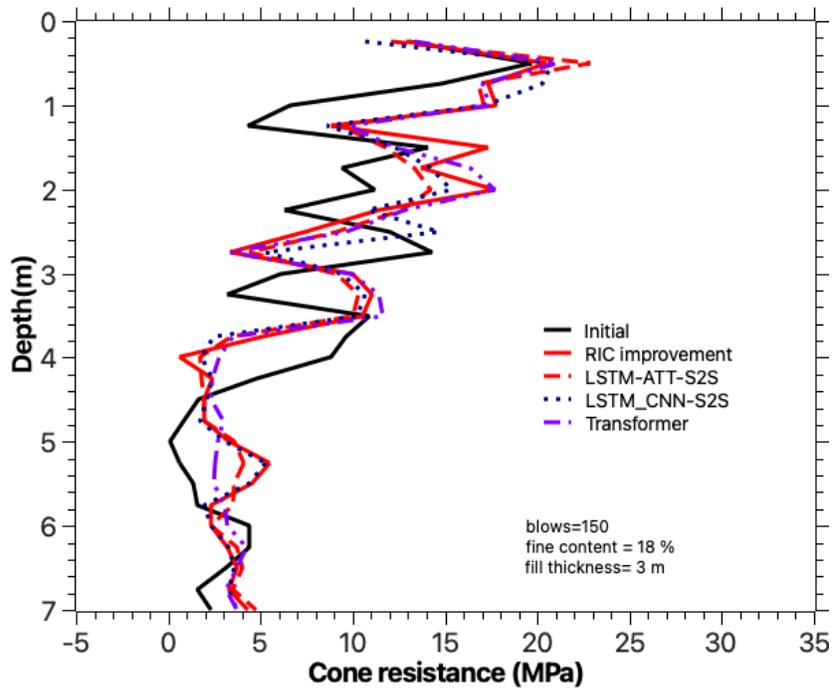
b) blows =150, fine content =18 % and fill thickness=3 m

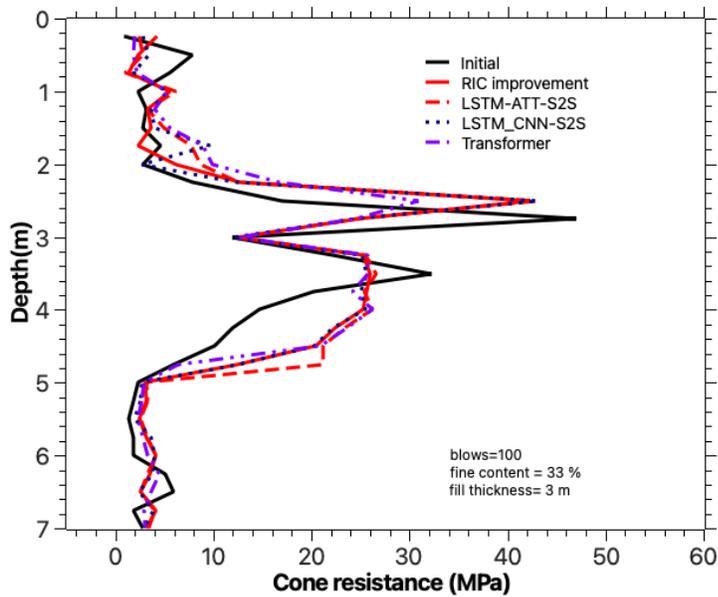
c) blows =100, fine content =33 % and fill thickness=3 m

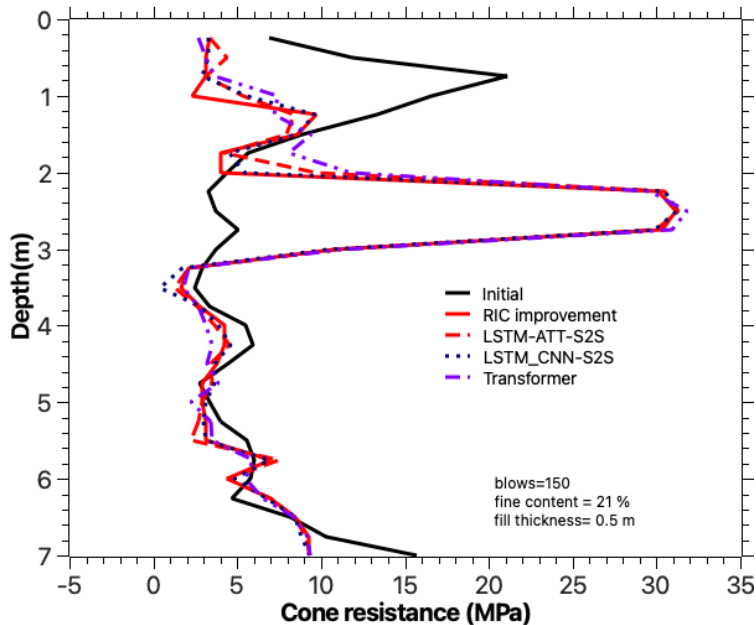
d) blows =150, fine content =21 % and fill thickness=0.5 m
**Figure. 22** Cone resistance of actual and predicted values for different blows, fine content and fill thickness using the actual previous sequential data as input value

### 3.5. Application of trained model
      The proposed model can generate the outcome of the RIC test as illustrated in Fig. 23 without using the previous sequential output data. The model can generate the output step by step using its predicted value as the input for the next sequence time step. This is the same architecture of the generative large language model. The model has an encoder-decoder architecture that can generate the output step by step, using its own predictions as inputs for the next time steps. The model works as follows:
- The initial soil profile is input and embedded with a feature vector to form a combined tensor.

- The combined tensor is fed into the encoder, which encodes the soil profile into a latent representation.
- The initial post-improvement profile is represented by a one-dimensional tensor with zero values.
- The decoder takes the latent representation and the initial post-improvement profile as inputs and predicts the cone resistance value at the second depth.
- The predicted value is appended to the post-improvement profile vector, which is then fed back into the decoder along with the latent representation to predict the next depth. This process is repeated until the desired depth (7 m) is reached.

This architecture can reduce the computation cost, as the encoder only needs to encode the initial soil profile once, while the decoder iterates until the desired depth is reached.

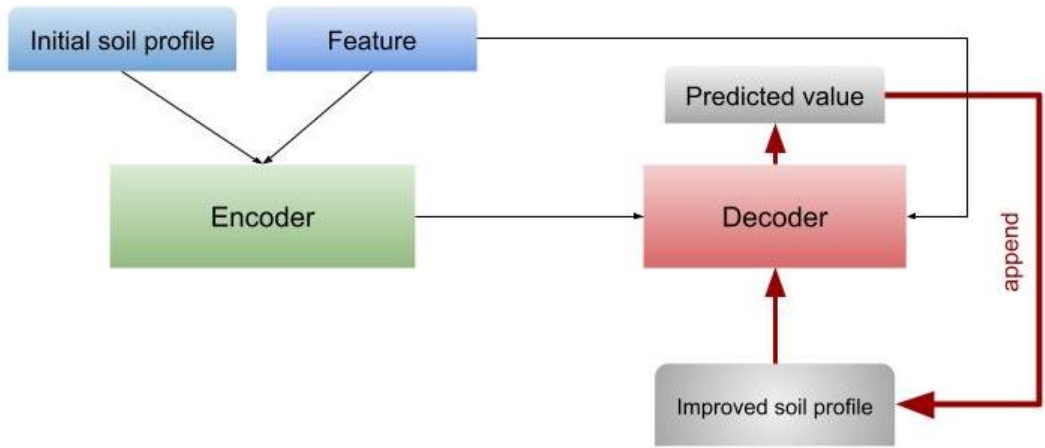

**Figure 23** The architecture of the prediction of generative model

The proposed model can generate the outcome of the RIC test as illustrated in Fig. 23 without using the previous sequential output data. The model can generate the output step by step using its predicted value as the input for the next sequence time step. This is the same architecture of the generative large language model. Schematic diagram of transformer model

The generative model can simulate the outcome of RIC from only the initial values of soil profile and compaction features as inputs. The model produces the post-compaction results of RIC at every 0.25 m interval (Fig. 24). The model performs better at the shallow depth than the deeper layer in generating the cone resistance, due to the cumulative error from the generative process. The model generates the output sequentially from the ground surface to the deeper ground. If the results from the shallow depth have some error, it might propagate to a higher error for the prediction in the deeper layer. This is because sequence to sequence models are trained using teacher forcing, which means that they use the ground truth tokens as inputs for the next time step during training. However, during inference, they use their own predictions as inputs for the next time step. This can lead to a mismatch between the training and inference distributions and cause the model to generate erroneous or repetitive tokens when the output sequence is long (Bengio et al. 2015). Another reason might be related to the long sequence of LSTM layer in the decoder section (28 sequences). It might suffer from the vanishing or exploding gradient problem in LSTM network (Ahmadvand et al. 2019).

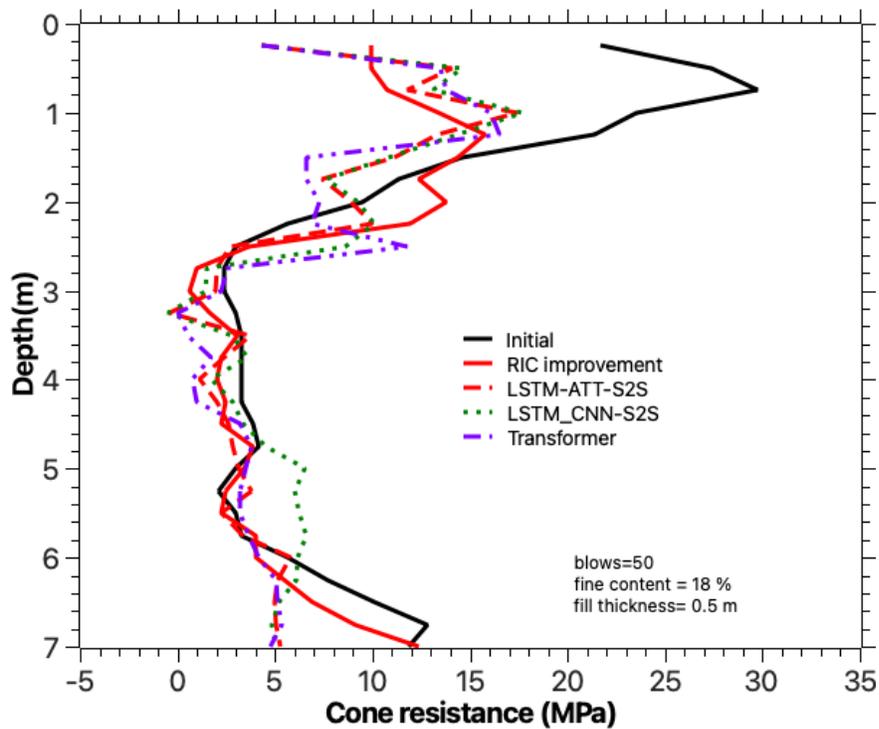

**Figure. 24** Cone resistance of actual and predicted values from generative AI with different model architecture

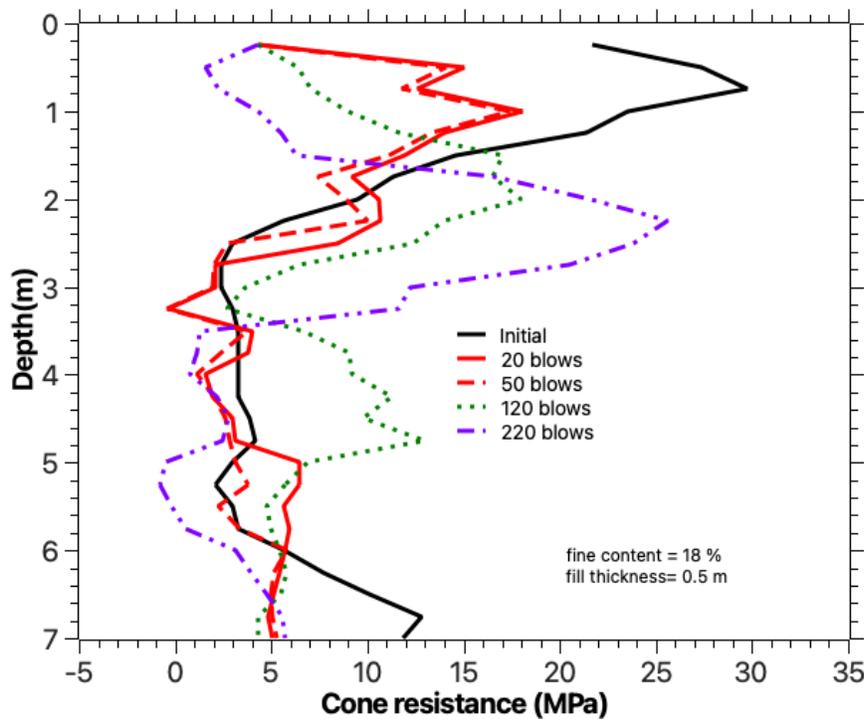

**Figure. 25** The predicted outcome of RIC with different applied energy (blows) from generative AI

The parametric study was conducted by using the trained model, LSTM-ATT-S2S, with varying energy levels for the compaction using RIC (Fig. 25). The input soil profile was based on a real case with a fill thickness of 0.5 m and a fine content of 18%. The number of blows was not similar to the actual field test to demonstrate the performance of the generative model. The predicted cone resistance increased with increasing hammer blows of RIC. The maximum cone resistance occurred at the depth of 2.5 m and tended to be located at the deeper layer with higher compaction effort. The prediction also showed the cone resistance with blow = 20, which was approximately similar to the blow = 50 case. However, there was an anomalous result of blow = 120 at the depth of 4 m, where the cone resistance increased up to about 10 MPa. This might be due to the training data from another location having a high cone resistance at this depth, which influenced the model weight and bias. However, for higher blows, the cone resistance tended to reduce to the initial value. This indicated that more test data was required to train the model and improve its efficiency in generating the output.

We developed a generative model based on a sequence-to-sequence architecture to simulate the effects of Rapid Impact Compaction (RIC) on various soil profiles. The model accurately predicted the trend of ground improvement by RIC and could assist engineers in estimating the expected outcomes before performing the actual compaction. Furthermore, the model could integrate more testing data from the site during the construction process, which could improve its efficiency and reliability. The model could also be trained in real time during the ground improvement construction. The sequence-to-sequence architecture of the model was also suitable for other geotechnical engineering problems, such as simulating the load-settlement curve of pile, bearing capacity, settlement, etc. We encoded the soil profile from different field tests, such as Standard Penetration Test (SPT), and added other soil parameters, such as soil type, strength and index properties. The encoded data served as input to train the model to simulate the geotechnical engineering outcomes.

## 4. Conclusion

In this study, we developed a transformer generative model to predict the outcome of rapid impact compaction (RIC) tests. We performed feature engineering to examine the influence of various features on the RIC results. We also evaluated different types of deep learning models for simulating the RIC outcome. The main findings and contributions of this study are summarized below:

- In this study, the association between the features and the RIC outcome was quantified using mutual information. The results showed that the initial cone resistance of soil was the most influential feature. The other feature, fill thickness, fine content, and blows of RIC hammer, has low influence compared to cone resistance, and they have similar influence to the improvement outcome.
- The effects of fill characteristics, compaction effort, and fill layer thickness on the compaction efficiency from RIC were also examined. The findings revealed that the compaction efficiency increased with lower fine content of fill, higher compaction effort, and thinner fill layer. However, there was a limit to the improvement of compaction efficiency when the compaction effort exceeded 100. Moreover, a decrease in cone resistance was detected when the compaction effort was above 100, which indicated a punching shear failure.
- Different types of feed forward model was access to find applicability to simulated the outcome of RIC. The more complex model can reduce the RMSE of the prediction, but it shows obviously overfit for the test data, especially using LSTM and CNN as deep learning model.
- Sequence to sequence architecture exhibit the good performance to predicted the outcome of RIC. The propose a triple hybrid model between LSTM, CNN and attention mechanism demonstrated the best performance among other type of model including original transformer model to simulate the outcome of RIC.
- The results of RIC were compared with the field test results and found a reasonable agreement between them, especially for the shallow depth.
- The application of using the trained mode as a generative mode was presented. It show reasonably performance to predict the unforeseen outcome of rapid impact compaction with different compaction effort.

## 5. References

Ahmadvand, S., M. Gharachorloo, and B. Minaei-Bidgoli. 2019. "A Survey on Natural Language Generation Techniques with a Focus on Dialogue Systems." *Journal of Artificial Intelligence and Data Mining*, 7 (2): 149–161.

Alhussein, M., K. Aurangzeb, and S. I. Haider. 2020. "Hybrid CNN-LSTM Model for Short-Term Individual Household Load Forecasting." *IEEE Access*, 8: 180544–180557. https://doi.org/10.1109/ACCESS.2020.3028281.

Bengio, S., O. Vinyals, N. Jaitly, and N. Shazeer. 2015. "Scheduled Sampling for Sequence Prediction with Recurrent Neural Networks." *Advances in Neural Information Processing Systems*, 1171–1179.

Born, J., and M. Manica. 2022. "Regression Transformer: Concurrent sequence regression and generation for molecular language modeling." arXiv. https://doi.org/10.48550/ARXIV.2202.01338.

Born, J., and M. Manica. 2023. "Regression Transformer enables concurrent sequence regression and generation for molecular language modelling." *Nature Machine Intelligence*, 5 (4): 432–444. https://doi.org/10.1038/s42256-023-00639-z.

Bradbury, J., R. Frostig, P. Hawkins, M. J. Johnson, C. Leary, D. Maclaurin, G. Necula, A. Paszke, J. V. Plas, S. W.- Milne, and Q. Zhang. 2018. "JAX: composable transformations of Python+ Num Py programs."

Chen, J., X. Li, and Z. Wang. 2019. "Sigmoid function: A brief introduction." *Journal of Physics: Conference Series*, 1168 (2): 022022. IOP Publishing.

Cheng, S.-H., S.-S. Chen, and L. Ge. 2021. "Method of estimating the effective zone induced by rapid impact compaction." *Scientific Reports*, 11 (1): 18336. Nature Publishing Group UK London.

Chollet, F. and others. 2015. "Keras."

Devlin, J., M.-W. Chang, K. Lee, and K. Toutanova. 2019. "BERT: Pre-training of Deep Bidirectional Transformers for Language Understanding." *North American Chapter of the Association for Computational Linguistics*.

Ding, M., Z. Yang, W. Hong, W. Zheng, C. Zhou, D. Yin, J. Lin, X. Zou, Z. Shao, H. Yang, and others. 2021. "CogView: Mastering Text-to-Image Generation via Transformers." *arXiv preprint arXiv:2105.13290*. https://doi.org/10.48550/arXiv.2105.13290.

Fang, L., T. Zeng, C. Liu, L. Bo, W. Dong, and C. Chen. 2021. "Transformer-based Conditional Variational Autoencoder for Controllable Story Generation." *arXiv preprint arXiv:2101.00828*.

Fukushima, K. 1980. "Neocognitron: A self-organizing neural network model for a mechanism of pattern recognition unaffected by shift in position." *Biol. Cybernetics*, 36 (4): 193–202. https://doi.org/10.1007/BF00344251.

Ghanbari, E., and A. Hamidi. 2014. "Numerical modeling of rapid impact compaction in loose sands." *Geomechanics and Engineering*, 6: 487–502.

Hochreiter, S., and J. Schmidhuber. 1997. "Long short-term memory." *Neural computation*, 9 (8): 1735–1780. MIT Press.

Khatun, Mst. A., M. A. Yousuf, S. Ahmed, Md. Z. Uddin, S. A. Alyami, S. Al-Ashhab, H. F. Akhdar, A. Khan, A. Azad, and M. A. Moni. 2022. "Deep CNN-LSTM With Self-Attention Model for Human Activity Recognition Using Wearable Sensor." *IEEE Journal of Translational Engineering in Health and Medicine*, 10: 1–16. https://doi.org/10.1109/JTEHM.2022.3177710.

Kingma, D. P., and J. Ba. 2015. "Adam: A method for stochastic optimization." *3rd International Conference on Learning Representations, ICLR 2015*.

LeCun, Y., L. Bottou, Y. Bengio, and P. Haffner. 1998. "Gradient-based learning applied to document recognition." *Proceedings of the IEEE*, 86 (11): 2278–2324. IEEE.

Liu, L., X. Liu, J. Gao, W. Chen, and J. Han. 2020. "Understanding the Difficulty of Training Transformers." *Proceedings of the 2020 Conference on Empirical Methods in Natural Language Processing (EMNLP)*, 2566–2577.

Midjourney, A., B. Smith, and C. Jones. 2023. "A New Approach to AI." *Journal of Artificial Intelligence*, 12 (3): 45–67. https://doi.org/10.1145/1234567.1234568.

Mirza, M., and S. Osindero. 2014. "Conditional generative adversarial nets." *arXiv preprint arXiv:1411.1784*.

Mohammed, M. M., R. Hashim, and A. F. Salman. 2010. "Effective improvement depth for ground treated with rapid impact compaction." *Scientific Research and Essays*, 5 (20): 3236–3246.

Mohammed, M., H. Roslan, and S. Firas. 2013. "Assessment of rapid impact compaction in ground improvement from in-situ testing." *Journal of Central South University*, 20 (3): 786–790. Springer.

OpenAI. 2023. "GPT-4 Technical Report." arXiv. https://doi.org/10.48550/ARXIV.2303.08774.

Paszke, A., S. Gross, F. Massa, A. Lerer, J. Bradbury, G. Chanan, T. Killeen, Z. Lin, N. Gimelshein, L. Antiga, A. Desmaison, A. Kopf, E. Yang, Z. DeVito, M. Raison, A. Tejani, S. Chilamkurthy, B. Steiner, L. Fang, J. Bai, and S. Chintala. 2019. "PyTorch: An Imperative Style, High-Performance Deep Learning Library." *Advances in Neural Information Processing Systems 32*, 8024–8035. Curran Associates, Inc.

Peng, Y., J. Qi, and Y. Yuan. 2018. "Stable Diffusion: A Generalized Framework for Transfer Learning in Convolutional Neural Networks." *Proceedings of the 27th International Joint Conference on Artificial Intelligence*, 2470–2476.

Sagnika, S., B. S. P. Mishra, and S. K. Meher. 2021. "An attention-based CNN-LSTM model for subjectivity detection in opinion-mining." *Neural Computing and Applications*, 33 (24): 17425–17438. https://doi.org/10.1007/s00521-021-06328-5.

Serridge, C. J., and O. Synac. 2006. "Application of the Rapid Impact Compaction (RIC) technique for risk mitigation in problematic soils."

Simpson, L. A., S. T. Jang, C. E. Ronan, and L. M. Splitter. 2008. "Liquefaction potential mitigation using rapid impact compaction." *Geotechnical Earthquake Engineering and Soil Dynamics IV*, 1–10.

"sklearn.feature_selection.mutual_info_regression." n.d. *scikit-learn*. Accessed August 9, 2023. https://scikit-learn/stable/modules/generated/sklearn.feature_selection.mutual_info_regression.html.

Spyropoulos, E., B. A. Nawaz, and S. A. Wohaibi. 2020. "A Case Study on Soil Improvement with Rapid Impact Compaction (RIC)." *WJET*, 08 (04): 565–589. https://doi.org/10.4236/wjet.2020.84040.

Su, X., J. Li, and Z. Hua. 2022. "Transformer-Based Regression Network for Pansharpening Remote Sensing Images." *IEEE Transactions on Geoscience and Remote Sensing*, 60: 1–23. https://doi.org/10.1109/TGRS.2022.3152425.

Tarawneh, B., and M. Matraji. 2014. "Ground improvement using rapid impact compaction: case study in Dubai." *Građevinar*, 66 (11.): 1007–1014. Hrvatski savez građevinskih inženjera.

Team, K. n.d. "Keras documentation: PositionEmbedding layer." Accessed August 16, 2023. https://keras.io/api/keras_nlp/modeling_layers/position_embedding/.

TensorFlow Developers. 2023. "TensorFlow." Zenodo.

Touvron, H., T. Lavril, G. Izacard, X. Martinet, M.-A. Lachaux, T. Lacroix, B. Rozière, N. Goyal, E. Hambro, F. Azhar, A. Rodriguez, A. Joulin, E. Grave, and G. Lample. 2023. "LLaMA: Open and Efficient Foundation Language Models." arXiv. https://doi.org/10.48550/ARXIV.2302.13971.

Van Houdt, G., C. Mosquera, and G. Nápoles. 2020. "A review on the long short-term memory model." *Artif Intell Rev*, 53 (8): 5929–5955. https://doi.org/10.1007/s10462-020-09838-1.

Vaswani, A., N. Shazeer, N. Parmar, J. Uszkoreit, L. Jones, A. N. Gomez, L. Kaiser, and I. Polosukhin. 2017. "Attention Is All You Need." arXiv. https://doi.org/10.48550/ARXIV.1706.03762.

Vukadin, V. 2013. "The improvement of the loosely deposited sands and silts with the Rapid Impact Compaction technique on Brežice test sites." *Engineering Geology*, 160: 69–80. Elsevier.

Wang, J., Z. Yang, X. Hu, L. Li, K. Lin, Z. Gan, Z. Liu, C. Liu, L. Wang, and others. 2021a. "GIT: A Generative Image-to-text Transformer for Vision and Language." *arXiv preprint arXiv:2205.14100*. https://doi.org/10.48550/arXiv.2205.14100.

Wang, Y., Y. Yang, J. Bai, M. Zhang, J. Bai, J. Yu, C. Zhang, G. Huang, and Y. Tong. 2021b. "Evolving Attention with Residual Convolutions." *arXiv preprint arXiv:2102.12895*.

Wei, Y., X. Liang, Z. Shen, and D. N. T. Huynh. 2021. "Unifying Multimodal Transformer for Bi-directional Image and Text Generation." *arXiv preprint arXiv:2110.09753*.

Youwai, S., S. Detcheewa, W. Kongkitkul, S. Suparp, and A. Sirisonth. 2023. "A Field Prototype Test of Rapid Impact Compaction for Ground Improvement and Backfill Compaction at U-Tapao Airport." *Proceeding of the 21st Southeast Asian Geotechnical Conference and 4th AGSSEA Conference*, (in press). Bangkok Thailand.

Zohourianshahzadi, Z., and J. K. Kalita. 2021. "Neural Attention for Image Captioning: Review of Outstanding Methods." *arXiv preprint arXiv:2111.15015*.

**List of Symbols**

| | |
|---|---|
| *X,Y* | input variable |
| *p(x,y)* | joint and marginal probability distribution |
| *p(x) p(y)* | probability distribution of *x* |
| *p(y)* | probability distribution of *y* |
| $MI(X;Y)$ | mutual information of X and Y |
| $E_{RIC}$ | efficiency of RIC |
| $q_c^{improve}$ | cone resistance after RIC |
| $q_c^{ini}$ | cone resistance before RIC |
| T | Fill thickness |
| F | Fine content |
| MAPE | mean absolute percentage error |
| $Y_{pred}$ | predicted value |
| $Y_{actual}$ | actual value |
| N | number of sample |
| Q | matrix of query |
| K | matrix of key |
| V | matrix of value |
| $d_k$ | dimension of key vector |
| pos | position |
| i | dimension |
| d_model | model dimension |
| *X'* | *Scaled data* |
| *X* | *unscaled data* |
| *μ* | *Mean value* |
| *σ* | *Standard deviation* |